\documentclass[twocolumn,secnumarabic,amssymb, nobibnotes, aps, pre]{revtex4-2}

\usepackage{natbib}  
\usepackage{caption} 
\usepackage{graphicx}
\usepackage{amssymb}
\usepackage{color}
\usepackage{amsmath}
\usepackage{epsfig}
\usepackage{multirow, makecell}
\usepackage{subcaption}
\usepackage{booktabs}
\usepackage{bm}
\usepackage{tabulary}
\usepackage{hhline}
\usepackage{multirow, rotating}
\usepackage{threeparttable}
\usepackage{amsfonts}
\usepackage{hyperref} 

\newcommand{\Transpose}{^{\mathsf{T}}}

\newcommand{\HiddenDim}[1]{n_{#1}}

\newcommand{\LeftSingularVectorSymb}{S}
\newcommand{\LeftSingularVector}{\LeftSingularVectorSymb}

\newcommand{\RightSingularVectorSymb}{D}
\newcommand{\RightSingularVector}{\RightSingularVectorSymb}

\newcommand{\SingularValueSymb}{V}
\newcommand{\SingularValue}{\SingularValueSymb}

\begin{document}

\preprint{APS/123-QED}

\title{
Noise-robust Contrastive Learning for Critical Transition Detection in Dynamical Systems
}

\author{Wenqi Fang$^{1}$}
\email[]{wq.fang@siat.ac.cn}
\author{Ye Li$^{1}$}
\affiliation{$^1$Shenzhen Institutes of Advanced Technology, Chinese Academy of Sciences}





\date{\today}

\begin{abstract}
Detecting critical transitions in complex, noisy time-series data is a fundamental challenge across science and engineering. Such transitions may be anticipated by the emergence of a low-dimensional order parameter, whose signature is often masked by high-amplitude stochastic variability. Standard contrastive learning approaches based on deep neural networks, while promising for detecting critical transitions, are often overparameterized and sensitive to irrelevant noise, leading to inaccurate identification of critical points. To address these limitations, we propose a neural network architecture, constructed using singular value decomposition technique, together with a strictly semi-orthogonality-constrained training algorithm, to enhance the performance of traditional contrastive learning. Extensive experiments demonstrate that the proposed method matches the performance of traditional contrastive learning techniques in identifying critical transitions, yet is considerably more lightweight and markedly more resistant to noise.

\end{abstract}

\maketitle


\section{Introduction}
Critical transitions \cite{PhysRevX.14.031009}, also known as tipping points \cite{PhysRevResearch.6.043194} or phase transitions \cite{Ma_2023, Han_2023} in different contexts, are of broad significance across disciplines because they enable the anticipation of abrupt, nonlinear changes in system behavior. 
Many complex dynamical systems, including ecosystems \cite{flores2024critical}, climate systems \cite{rietkerk2025ambiguity}, financial markets \cite{de2025enhancing}, and biological or medical systems \cite{aihara2022dynamical, chen2025ultralow}, exhibit such behavior and may appear stable even as slowly evolving external or internal conditions drive them toward a critical threshold. However, once this threshold, often characterized as a bifurcation point \cite{chen2000bifurcation}, is crossed, the system may undergo a sudden, and sometimes irreversible, transition to a qualitatively different state. Such regime shifts may lead to severe outcomes, such as ecosystem collapse \cite{dylewsky2023universal}, economic disruptions \cite{perez2024using}, and public health challenges \cite{deb2022identifying}. Consequently, the capability to detect and anticipate critical transitions in these systems is crucial for effective risk mitigation and loss prevention.

Motivated by this need, numerous early warning signals (EWSs) have been proposed over the past few decades to forecast impending critical transitions \cite{scheffer2009early, quansah2010early, scheffer2012anticipating}. In general, these approaches are grounded in the concept of critical slowing down (CSD), whereby a dynamical system approaching a tipping point or bifurcation exhibits progressively slower recovery from small perturbations. Such reduced resilience is typically manifested as increases in lag-1 autocorrelation, variance, or Pearson correlation coefficients \cite{scheffer2012anticipating, dakos2008slowing, aihara2022dynamical, ma2025predicting}. A representative example is the dynamical network biomarkers (DNB) framework, which extends the concept of CSD to high-dimensional systems with complex network structures and has been widely applied to the detection of pre-disease states \cite{aihara2022dynamical}. Beyond CSD-based methods, the rapid development of neural networks (NN) has fueled growing interest in deep learning (DL)–based approaches, including DL classifiers \cite{bury2021deep, ma2025predicting}, graph isomorphism network methods \cite{PhysRevX.14.031009}, modified reservoir computing schemes \cite{kong2021machine, panahi2024machine}, and contrastive learning (CL)–based frameworks \cite{gokmen2021symmetries, han2023framework, cy2023studying, romeo2025characterizing}. Building on these methods and informed by a careful assessment of the CL approach, this paper seeks to refine this idea in the following two key aspects.

Firstly, existing CL approaches typically rely on dense multilayer perceptrons (MLPs) to extract latent representations from the input data. Nevertheless, these fully connected (FC) layers commonly used in DL models are often substantially overparameterized \cite{yang2025random, alon2025does, xu2025overview}, raising the question of whether a more compact architecture can still capture the essential features associated with critical transitions. Building on this motivation, and inspired by the recent Meta-COMET framework \cite{fang2025constants}, we introduce a singular value decomposition (SVD)-based matrix factorization to parameterize the NN weights, thereby significantly reducing the number of trainable parameters. This approach is closely related to the low-rank modeling techniques described in \cite{cherukuri2025low, borsoi2025low} as well as to reduced-order modeling methods \cite{barraza2024reduced, corrochano2025predictive}. Furthermore, to support this SVD-structured architecture, we develop a training algorithm that enforces exact semi-orthogonal constraints on the factorized weight matrices, in contrast to the two-phase training procedure in \cite{fang2025constants}, which imposes semi-orthogonality only via a soft regularization term.

Secondly, in the literature on CL for detecting critical transitions, robustness to noise has received comparatively limited attention, despite the extensive development of robust CL methods for handling label noise, such as sample-selection strategies \cite{ortego2021multi, li2022selective} and symmetric loss formulations \cite{chuang2022robust, cui2025inclusive}. Moreover, many EWS-based approaches can fail to signal impending transitions when time-series data are noisy, as input perturbations may obscure meaningful signals and induce spurious sudden jumps \cite{ma2025predicting}. Thus, developing a noise-robust CL framework for critical transition detection is essential. To this end, and unlike existing robust CL strategies, we employ a SVD-based formulation, which is well known for its noise-attenuating properties \cite{sadek2012svd, epps2019singular}, coupled with a training algorithm that strictly enforces orthogonality constraints. Across a suite of representative dynamical systems, our experimental results further demonstrate that the proposed approach achieves performance comparable to, or even exceeding, that of FC–based CL models in detecting critical transitions under noisy conditions.

Based on the above discussion, in this paper, we propose a CL–based framework, termed SVDCL, for detecting critical transitions in dynamical systems. Our results show that SVDCL is both more lightweight and more robust to noise than traditional CL-based approach. The main contributions of this paper are threefold:

1. Taking inspiration from Meta-COMET \cite{fang2025constants}, we introduce an SVD-enhanced neural architecture that substantially compresses FC-based CL models, reducing the parameter count to only 80\% of the original in the SNIChop system, while preserving overall performance.

2. Unlike conventional CL training procedures, SVDCL employs a training algorithm that strictly enforces semi-orthogonality, specifically tailored to the SVD-based network architecture. This algorithm ensures that the corresponding weight matrices rigorously satisfy the semi-orthogonal constraints inherent to the SVD formulation.

3. We validate SVDCL across a diverse set of dynamical systems, ranging from simple models such as SNIChop to more complex systems like the quantum Ising model. Remarkably, despite its lightweight architecture, SVDCL achieves performance comparable to, or even surpasses, conventional CL models in detecting critical transitions under noisy conditions.

\section{Methodology}
The overall framework of our SVDCL methodology is shown in Figure~\ref{fig:framework}. To detect critical transitions, the collected inputs $x$, which are trajectories or spin configurations in our experiments, are processed through an
$L$-layer SVD-based NN to produce the latent feature $h^{L+1}$. 
Meanwhile, the augmented input $\tilde{x}$, obtained by transforming the original $x$, is fed into the same NN to generate $\tilde{h}^{L+1}$. During training, we employ the standard InfoNCE contrastive loss $\mathcal{L}_{infoNCE}$ \cite{chen2020simple}. In the testing stage, critical transitions are detected using metrics such as similarity, variance, and mutual similarity \cite{aihara2022dynamical, gokmen2021symmetries, han2023framework, cy2023studying, romeo2025characterizing}.

\subsection{SVD-based neural network architecture}

\label{sec:framework}
\begin{figure*}[t]
    \centering
    \includegraphics[width=\linewidth]{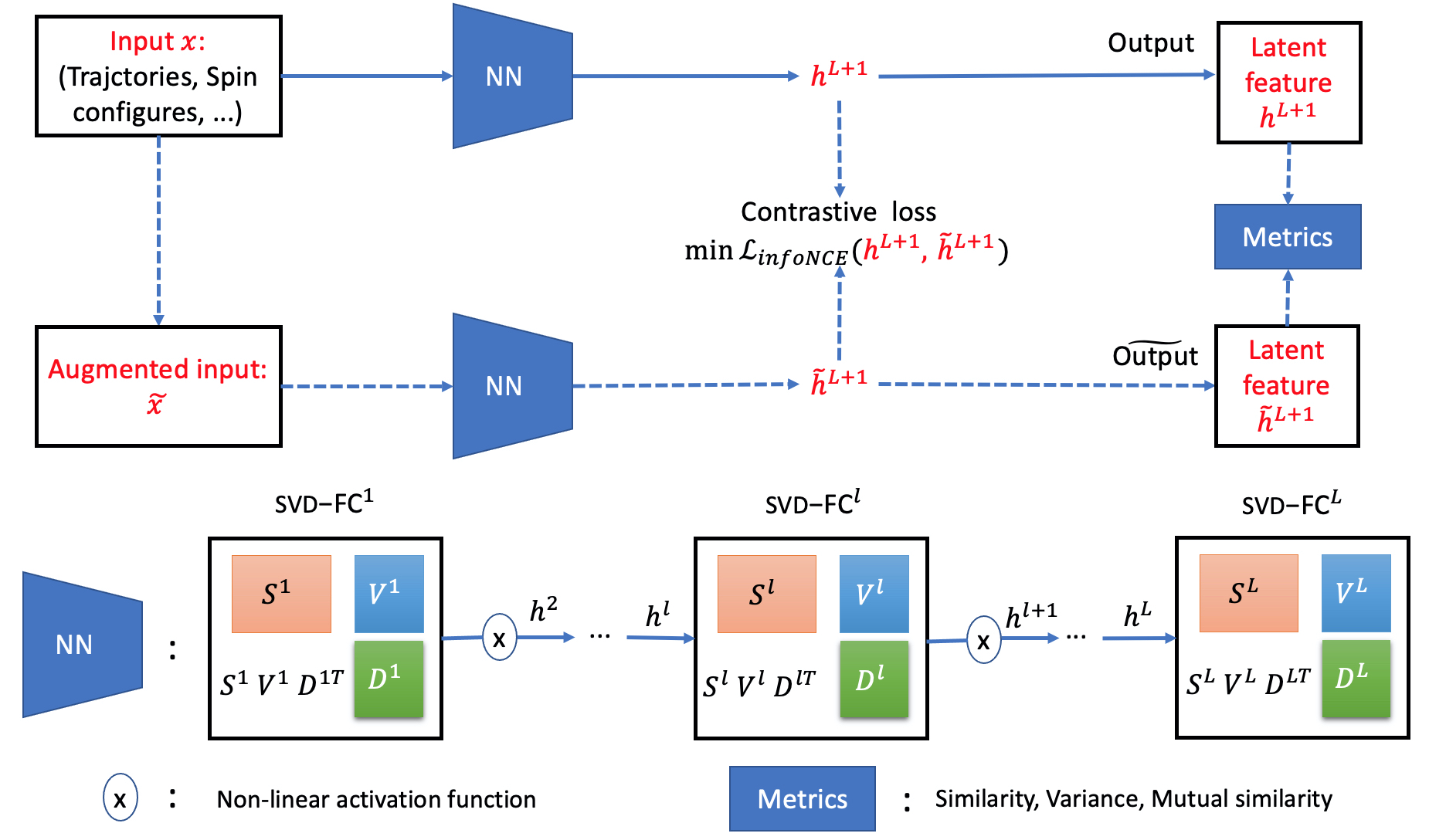}
    \caption{Overall of our SVDCL framework.
}
\label{fig:framework}
\end{figure*}

Previous CL-based approaches for critical transition detection typically rely on FC layers to encode latent features of the underlying dynamical systems \cite{gokmen2021symmetries, han2023framework, cy2023studying, romeo2025characterizing}. However, this design inherently introduces redundant parameters \cite{yang2025random, alon2025does, xu2025overview} and offers limited robustness to noise \cite{cooper2018loss, adeoye2024regularized}.

To overcome these limitations, we resort to the SVD trick to construct our NN architecture, shown in Figure~\ref{fig:framework}. As illustrated in Figure~\ref{fig:framework}, each layer, denoted SVD-FC$^l$, consists of three components-$S^l$, $V^l$ and $D^l$-with all parameters distinct across layers. The mapping between consecutive two layers, $l$ and $l+1$, is given by: 
\begin{align}\label{eq:factored_fc}
    &  h^{l+1} = f(\textsc{svd-FC}^{l}(h^{l})) \  \Leftrightarrow \ h^{l+1} = f(\LeftSingularVector^l(\SingularValue^l (\RightSingularVector^l{}\Transpose h^l))),
\end{align}
where $\LeftSingularVector^l \in \mathbb{R}^{\HiddenDim{l+1} \times r}$ and $\RightSingularVector^l \in \mathbb{R}^{\HiddenDim{l} \times r}$ are full column-rank matrices (i.e., rank $r \ll n_l, n_{l+1}$), $\SingularValue^l \in \mathbb{R}^{r \times r}$ is a diagonal matrix and $f$ is a non-linear activation function. With this SVD-based low-rank reparameterization, the number of tunable parameters can be reduced substantially. Furthermore, by constraining the weight matrix to this low-rank structure, the SVD-based NN effectively functions as a nonlinear denoising filter, preferentially capturing the persistent, low-dimensional dynamics while suppressing high-variance noise. In contrast to unconstrained FC layers, the semi-orthogonal structure imposed via SVD decomposition facilitates stable and independent feature extraction, yielding representations that are both more robust and better suited for detecting critical transitions \cite{fang2025constants}.

Both the original input $x$ and its augmented version $\tilde{x}$ 
are processed by the same SVD-based NN.
The internal computations for obtaining $h^{L+1}$ ($\tilde{h}^{L+1}$), starting from $h^{1} = x$ ($\tilde{h}^{1} = \tilde{x}$), are described as follows:
\begin{equation}\label{aaa}
    h^{l+1} = f(S^l ( ReLU(V^l) (D^l{}\Transpose h^l))), l=1,\ldots,L,
\end{equation} 
where $f$ becomes an identity function when $l=L$. The ReLU activation truncates negative values, allowing $V^l$ 
to adaptively adjust its effective rank. This mechanism reduces the need to manually specify the rank $r$ for each layer.
According to equations~(\ref{aaa}), $h^{L+1}$ and $\tilde{h}^{L+1}$ are jointly determined once the layer widths $n_l$, rank $r$, and other relevant parameters are specified.

Overall, our architecture resembles the FC-based NN. However, for the same width and depth, the inclusion of the $\textsc{svd-FC}^{l}$ blocks substantially reduces the number of parameters, making our model significantly less redundant. Moreover, incorporating SVD techniques theoretically enhance the noise robustness of our architecture according to the earlier published work \cite{golub2013matrix, fang2025constants}. Both of these advantages are thoroughly validated in our experiments.

\subsection{Training scheme with strict semi-orthogonal constraints}

In conjunction with our SVD-based NN architecture, we develop a training algorithm that strictly enforces semi-orthogonal constraints on the parameters $S^l$ and $D^l$. This stands in contrast to the two-phase training method in \cite{fang2025constants}, which imposes only soft semi-orthogonal constraints through regularization terms, and thus aligns more effectively with our model structure.

To preserve the semi-orthogonality of $S^l$ and $D^l$ throughout optimization, a three-stage update procedure is applied at every optimization step. In this process, gradients are first computed for all three components ($S^l, V^l, D^l$), after which the semi-orthogonal constraints on $S^l$ and $D^l$ are explicitly enforced, and $V^l$ is truncated in accordance with the SVD definition. The detailed training strategy proceeds as follows:
\begin{itemize}
    \item The NN parameters ($S^l, V^l, D^l$) are updated using a standard optimizer (e.g., SGD or Adam \cite{sun2020optimization}) based on the computed contrastive loss $\mathcal{L}_{infoNCE}$:
    \begin{align}\label{bbb}
    &S^{l} = S^l - \eta \nabla_{S^l} \mathcal{L}_{infoNCE}, \\ \nonumber
    &V^{l} = V^l - \eta \nabla_{V^l} \mathcal{L}_{infoNCE}, \\ 
    &D^{l} = D^l - \eta \nabla_{D^l} \mathcal{L}_{infoNCE}.  \nonumber
    \end{align} 
    where $\eta$ denotes the learning rate, chosen to be identical for all the parameters. Notably, $V^l$ is theoretically required to remain diagonal during optimization. However, standard optimizers cannot preserve this property. Therefore, it is typically initialized as a low-dimensional vector and subsequently transformed into a diagonal matrix in the numerical implementation.
    \item After updating $S^l$ and $D^l$ via gradient descent, we apply $\mathcal{SVD}$ operation to each of them: 
    \begin{align}\label{ccc}
    S^{l}_{svd}, \ \_, \ Sh^{lT}_{svd} = \mathcal{SVD}(S^l)  \\ \nonumber
    D^{l}_{svd}, \ \_, \ Dh^{lT}_{svd} = \mathcal{SVD}(D^l).
    \end{align}
    Then, we project these matrices back onto the manifold of semi-orthogonal matrices by recombining their SVD components:
    \begin{align}\label{ddd}
    S^l = S^{l}_{svd}Sh^{lT}_{svd}\\ \nonumber
    D^l = D^{l}_{svd}Dh^{lT}_{svd}.
    \end{align}
    Thanks to the properties of the SVD, this step ensures that the columns of $S^l$ and $D^l$ remain orthonormal, thereby guaranteeing that the composed weight matrix $S^l V^l D^{lT}$ strictly satisfies the low-rank constraint and preserves its optimal denoising capabilities.
    \item To further stabilize training and preserve physical validity, the diagonal elements of $V^l$ (the singular values) are clipped to be non-negative using a ReLU activation, as stated in equation~\ref{aaa}. This guarantees $V_{ii} \geq 0$, consistent with its definition. Additionally, a fixed rank $r$ can also be imposed by keeping only the $r$ largest singular values and eliminating the smaller, less significant ones.
\end{itemize}
This iterative post-update orthogonalization constitutes the key procedural innovation that enables successful training of the SVDCL model, while fully exploiting the low-rank structural constraints inherent in the SVD to achieve effective representation learning and optimal denoising.

Owing to the proposed training strategy, our method reinforces the structural robustness of the SVD architecture. By strictly enforcing semi-orthogonality on $S^l$ and $D^l$ and truncating high-frequency singular components, the proposed SVDCL approach effectively performs a low-rank denoising projection. This projection preserves only the dominant, physically meaningful variations in the data while suppressing irrelevant noise. Consequently, the integration of our architectural design with the proposed training procedure yields a compact and stable subspace representation of the system dynamics, capturing their intrinsic structure while accentuating critical transitions for more reliable detection.

\section{Experiments}

We conduct extensive experiments on three classical dynamical system and one quantum spin model to illustrate the benefit of our SVDCL method. The details of the selected system are as follows:
\begin{itemize}
\item \textbf{SNIC-Hopf (SNIChopf):} 
A minimal model of nonreciprocal dynamics relevant to active matter and ecological systems, exhibiting three distinct behaviors separated by well-characterized bifurcations in the absence of noise \cite{martin2025transition}.

\item \textbf{Saddle-Homoclinic Orbit excitable system (SHO):} 
Similar to the SNIChopf system, this example represents a more complex parameterized system that undergoes both local and global bifurcations \cite{izhikevich2000neural}.



\item \textbf{Non-linear cell-cycle model (Cellcycle):}
A six-dimensional (D) cell division cycle model of cdc2–cyclin interaction, capturing key feedback mechanisms and exhibiting three distinct dynamical regimes \cite{tyson1991modeling}.

\item \textbf{2D Ising model (Ising):} 
A 2D quantum spin lattice model with a rigorously analyzed and well-characterized phase transition, serving as a canonical benchmark for studying critical phenomena and collective behavior \cite{onsager1944crystal}.

\end{itemize}
The analytical formulations of these models are described in recent references \cite{romeo2025characterizing, han2023framework}. Their corresponding ground-truth phase diagrams are provided in Appendix~\ref{ground truth}. To characterize the critical transition, three metrics are introduced:
\begin{itemize}
\item \textbf{Average similarity:} It is introduced to measure the cosine similarity between two adjacent normalized latent features with shape $B \times d$ ($B$: batchsize, $d$: feature dimension), extracted by NN, formulated as:
\begin{align}\label{ddd}
\text{Similarity} = \frac{1}{Bd}\sum F_i F^T_{i+1},
\end{align}
where $i$ represents the number of points along the control-parameter path (defined in Figure~\ref{fig:ground truth}). As the system approaches the critical point, feature similarity decreases sharply and exhibits amplified fluctuations, signaling the loss of configurational coherence and the growth of critical fluctuations.

\item \textbf{Variance:} Dual to the similarity metric, this quantity measures the variance of the latent features and is defined as:
\begin{align}
\text{Variance}
&= \sum_{j=1}^{d} Var\!\left(F, dim=0\right) \\  \nonumber
&= \sum_{j=1}^{d}
\frac{1}{B}
\sum_{i=1}^{B}
\left(
F_{ij}
- \frac{1}{B}\sum_{k=1}^{B} F_{kj}
\right)^{2}.
\end{align}
While feature similarity exhibits a sharp decrease at the critical point, feature variance exhibits a pronounced peak, reflecting the growth of critical fluctuations.

\item \textbf{Average mutual similarity:} This metric is defined in the same manner as the average similarity metric, except that it is computed pairwise for configurations sampled along the control-parameter path:
\begin{align}\label{ddd}
\text{Mutual Similarity} = \frac{1}{Bd}\sum F_i F^T_{j}.
\end{align}
The heatmap of mutual similarity evolves from a uniformly high-similarity pattern in the ordered phase to a fragmented, heterogeneous structure near the critical point, indicating the onset of a critical transition.
\end{itemize}

All the experiments were conducted using PyTorch on a remote server equipped with an NVIDIA A100 GPU (40 GB memory) and an Intel Xeon Gold 5320 CPU (26 cores). During the simulations, the training data were numerically generated and contaminated with additive Gaussian noise drawn from $\sqrt{2\sigma}\mathcal{N}(\mathbf{0}, \mathbf{I})$, while test data were independently generated and corrupted by Gaussian noise with the same noise level. Model training was conducted using PyTorch’s Adam optimizer combined with a cosine-annealing learning-rate schedule, and the SiLU activation function was used consistently across all experiments. To expedite training, an early-stopping criterion was triggered if the validation loss showed no improvement over 1000 (classical systems) or 300 (Ising) successive epochs. And we set $B=500$ in all the experiments. The benchmark method for comparison is a CL approach based on FC layers, referred to as MLPCL. Additional experimental details, including data preparation, data augmentation strategies, and NN settings, are provided in Appendix~\ref{experimental setting}. Our implementation and experiments are primarily based on the publicly available codebase \footnote{https://github.com/NicoRomeo/DynCarto}.

\subsection{SNIC-Hopf}
\begin{figure}[t]
    \centering
    \begin{minipage}[t]{0.48\columnwidth}
        \centering
        \includegraphics[width=\linewidth]{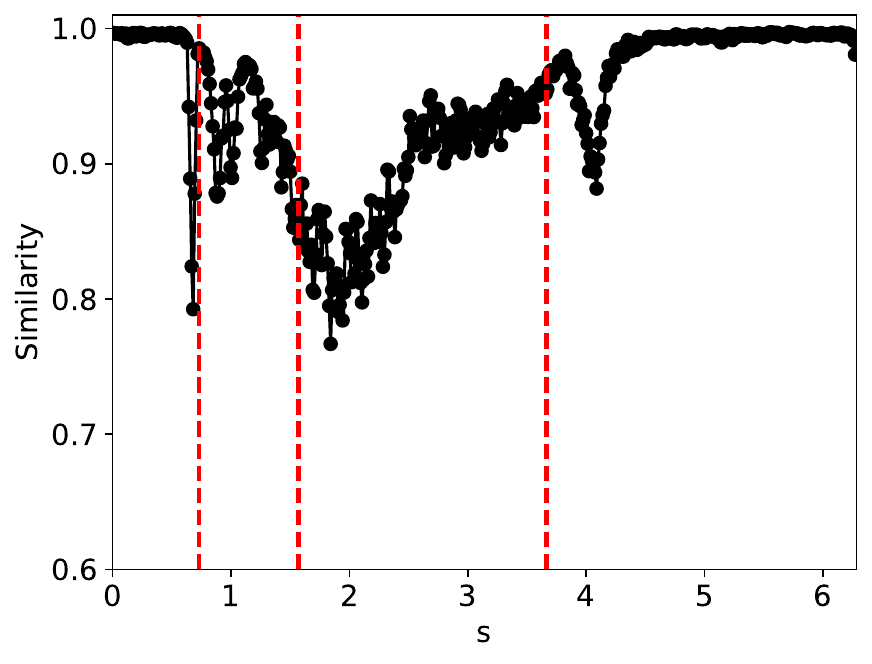}
    \end{minipage}
    \hfill
            \begin{minipage}[t]{0.48\columnwidth}
        \centering
        \includegraphics[width=\linewidth]{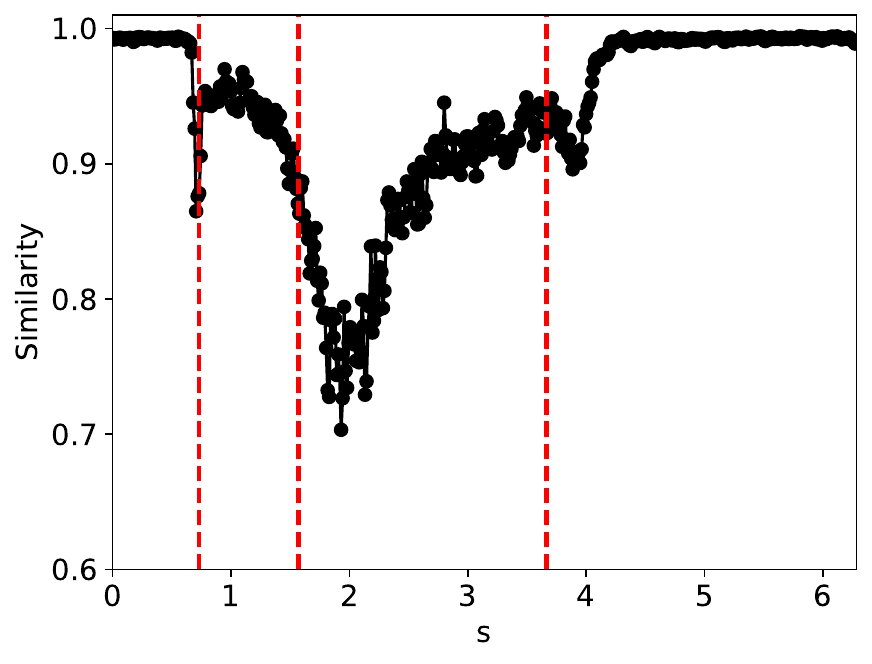}
    \end{minipage}
    \hfill
    \begin{minipage}[t]{0.48\columnwidth}
        \centering
        \includegraphics[width=\linewidth]{
        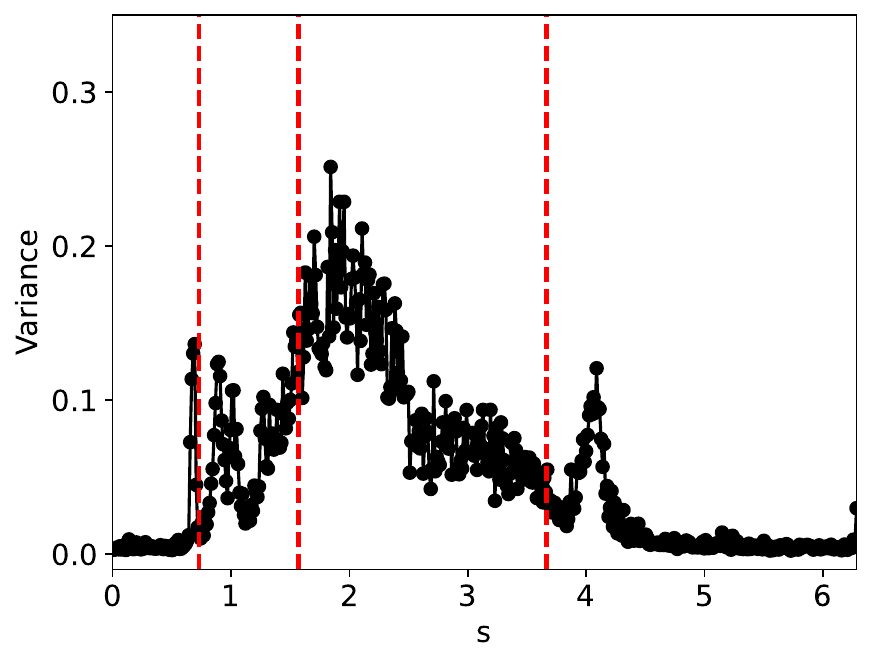}
    \end{minipage}
    \hfill
    \begin{minipage}[t]{0.48\columnwidth}
        \centering
        \includegraphics[width=\linewidth]{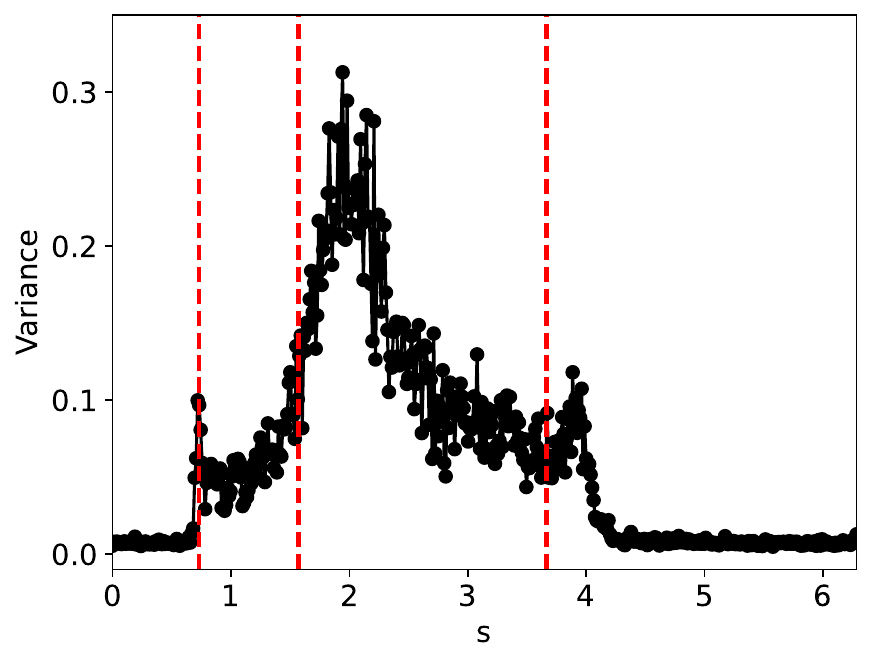}
    \end{minipage}
    \caption{Comparison result for SNIChopf system with $\sigma=0.005$: MLPCL (left column) and SVDCL (right column).}
    \label{fig:SNIChopf0.005}
\end{figure}

\begin{figure}[t]
    \centering
    \begin{minipage}[t]{0.48\columnwidth}
        \centering
        \includegraphics[width=\linewidth]{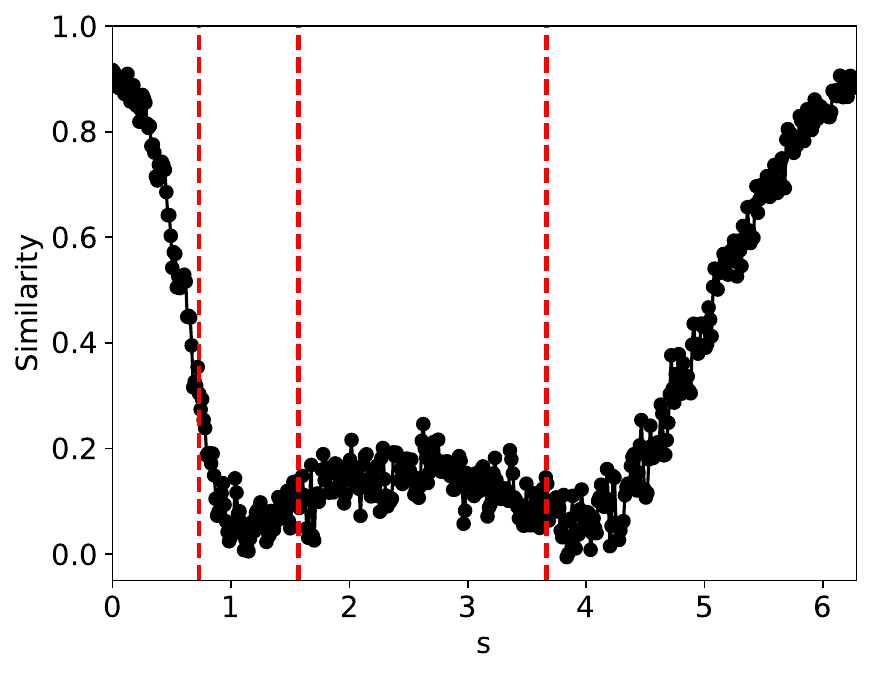}
    \end{minipage}
    \hfill
            \begin{minipage}[t]{0.48\columnwidth}
        \centering
        \includegraphics[width=\linewidth]{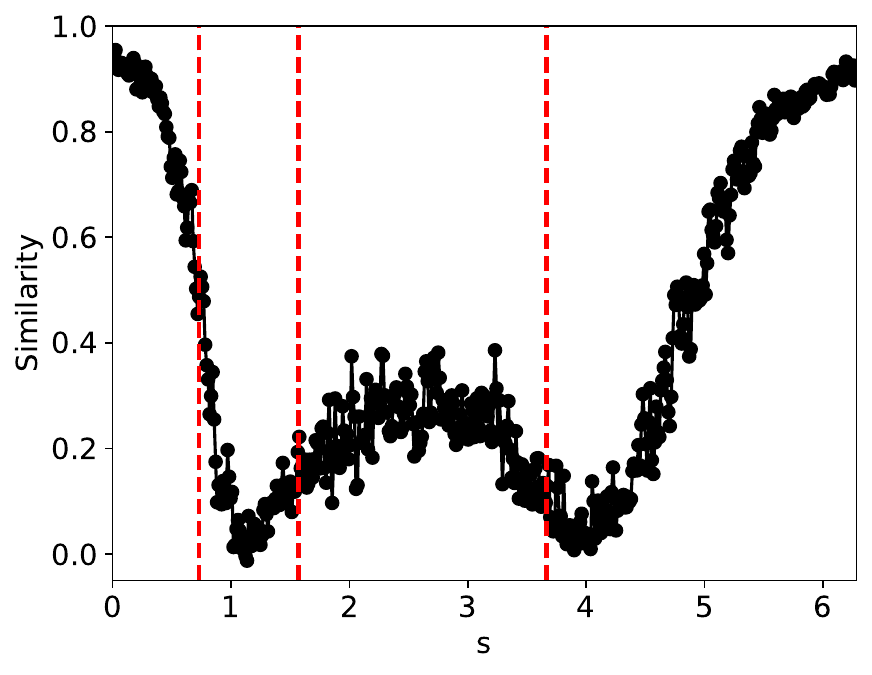}
    \end{minipage}
    \hfill
    \begin{minipage}[t]{0.48\columnwidth}
        \centering
        \includegraphics[width=\linewidth]{
        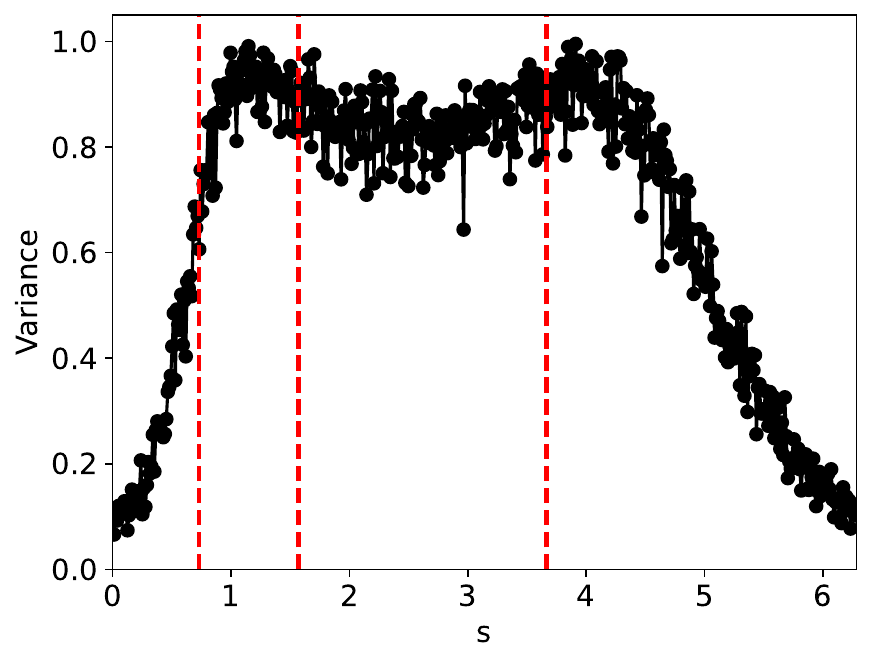}
    \end{minipage}
    \hfill
    \begin{minipage}[t]{0.48\columnwidth}
        \centering
        \includegraphics[width=\linewidth]{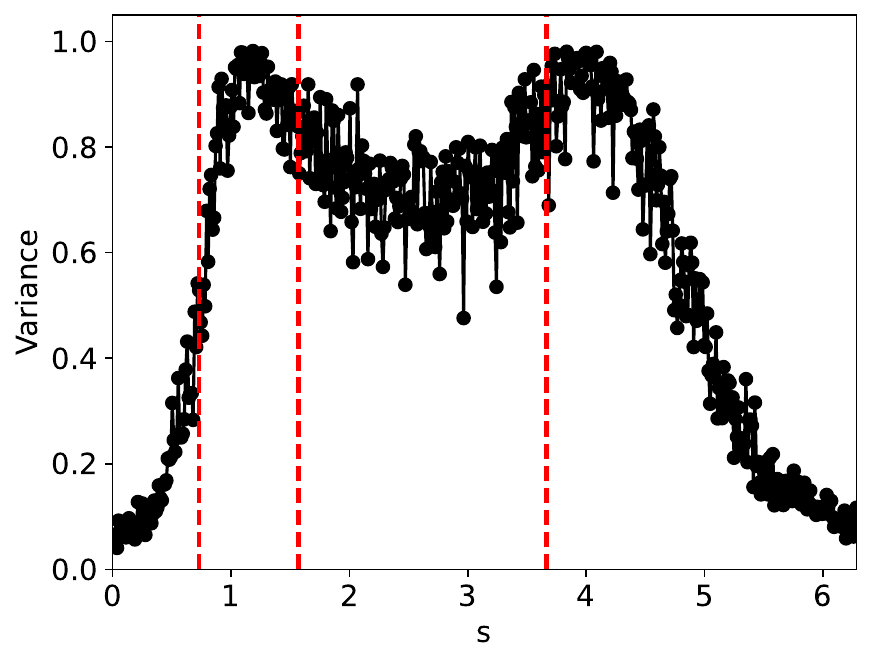}
    \end{minipage}
    \caption{The setting is identical to that of Figure~\ref{fig:SNIChopf0.005}, except with $\sigma = 0.39$.}
    \label{fig:SNIChopf0.39}
\end{figure}

The similarity and variance curves computed along the parameter path $(k(s), \alpha(s))$ (see Appendix~\ref{ground truth}) under different noise levels $\sigma$ (Figures~\ref{fig:SNIChopf0.005} and \ref{fig:SNIChopf0.39}) show that MLPCL and SVDCL exhibit closely aligned behaviors. This consistency indicates that both architectures successfully capture the same underlying phase structure, despite differing model constraints.

For the case $\sigma = 0.005$, the noise has a substantial impact on the system, as the resulting curves differ markedly from those obtained in the noise-free case ($\sigma = 0$; see Appendix~\ref{sec:classical}). As evidenced by the similarity and variance curves, both methods are still able to reliably approximate the location of the critical transition. However, in the vicinity of the critical point $s_1$, the SVDCL model demonstrates greater stability, exhibiting fewer spurious local minima or maxima. Moreover, near $s_2$, the metric curves produced by SVDCL exhibit a broader dynamic range, demonstrating its ability to preserve well-separated phase features in the latent space across different regions of the phase diagram, even under noisy conditions. While both methods fail to precisely capture the local extrema near $s_3$,  our approach shows comparatively better performance, yielding estimates that remain closer to the true extrema.
These phenomena underscore the robustness of our method, even when using only approximately 80\% of the original number of parameters.

When the noise level is increased to $\sigma = 0.39$, the overall patterns of the similarity and variance curves remain consistent with each other but differ markedly from those observed in the lower-noise cases, as shown in Figure~\ref{fig:SNIChopf0.39}. In this case, the results display only two well-defined local extrema, which approximately mark the true tipping points of the SNIChopf system, while no obviously distinct local structure appears at $s_2$. 
Nevertheless, SVDCL produces a wider similarity band around $s_2$, underscoring the discriminative strength of its low-rank–constrained latent space for phase-transition detection and demonstrating greater robustness to noise.

\subsection{Saddle-Homoclinic Orbit excitable system}
\begin{figure}[t]
    \centering
    \begin{minipage}[t]{0.48\columnwidth}
        \centering
        \includegraphics[width=\linewidth]{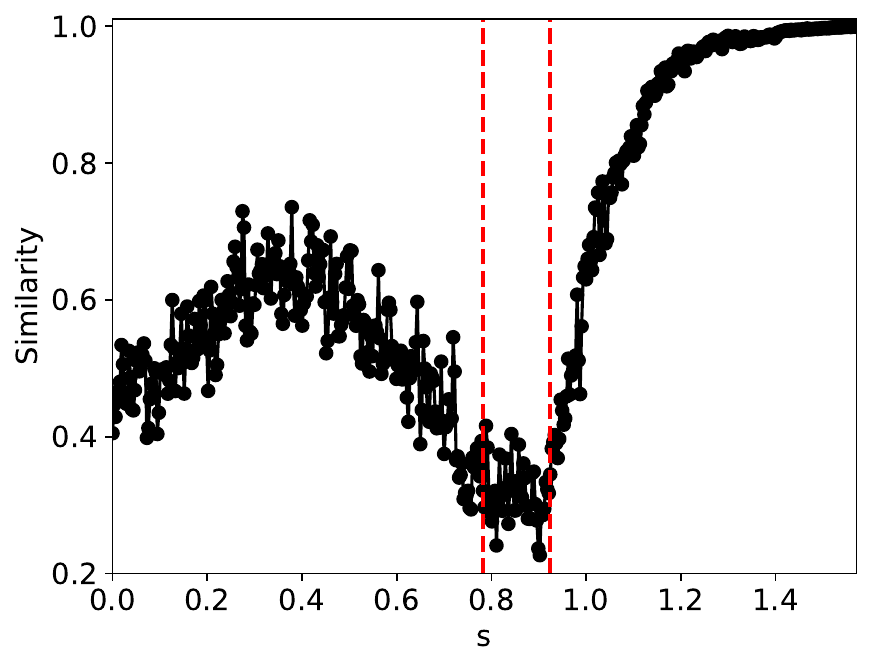}
    \end{minipage}
    \hfill
            \begin{minipage}[t]{0.48\columnwidth}
        \centering
        \includegraphics[width=\linewidth]{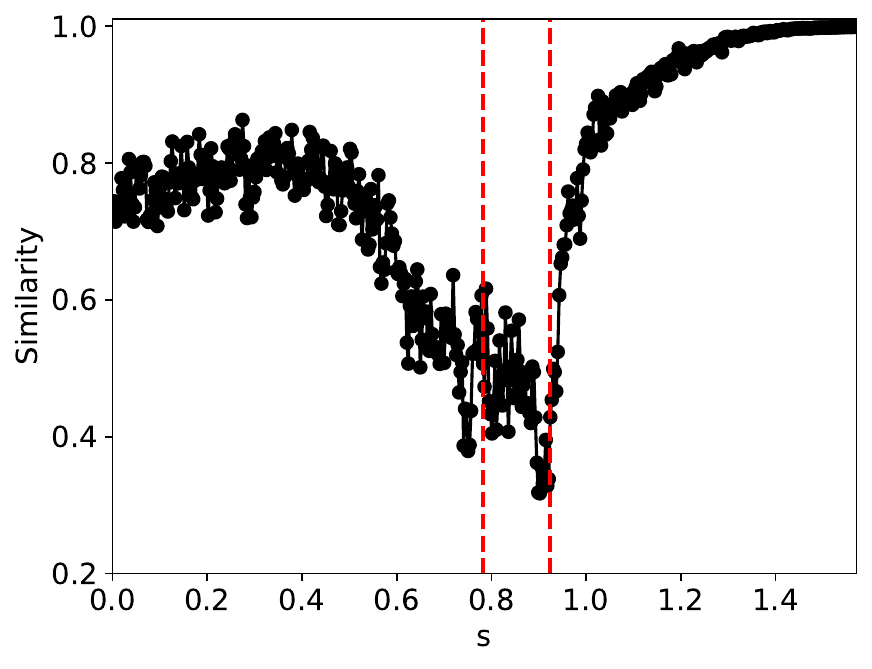}
    \end{minipage}
    \hfill
    \begin{minipage}[t]{0.48\columnwidth}
        \centering
        \includegraphics[width=\linewidth]{
        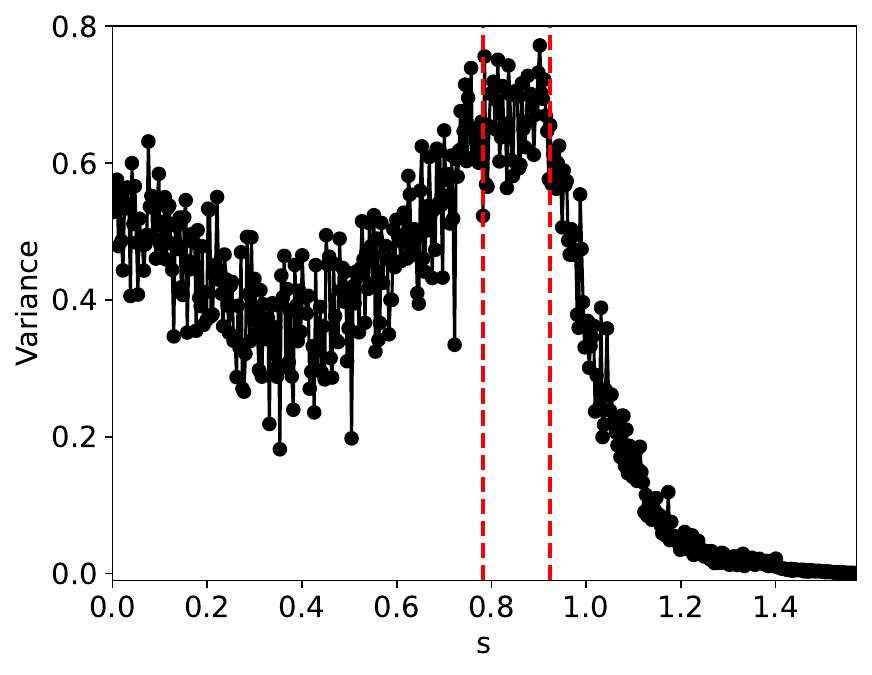}
    \end{minipage}
    \hfill
    \begin{minipage}[t]{0.48\columnwidth}
        \centering
        \includegraphics[width=\linewidth]{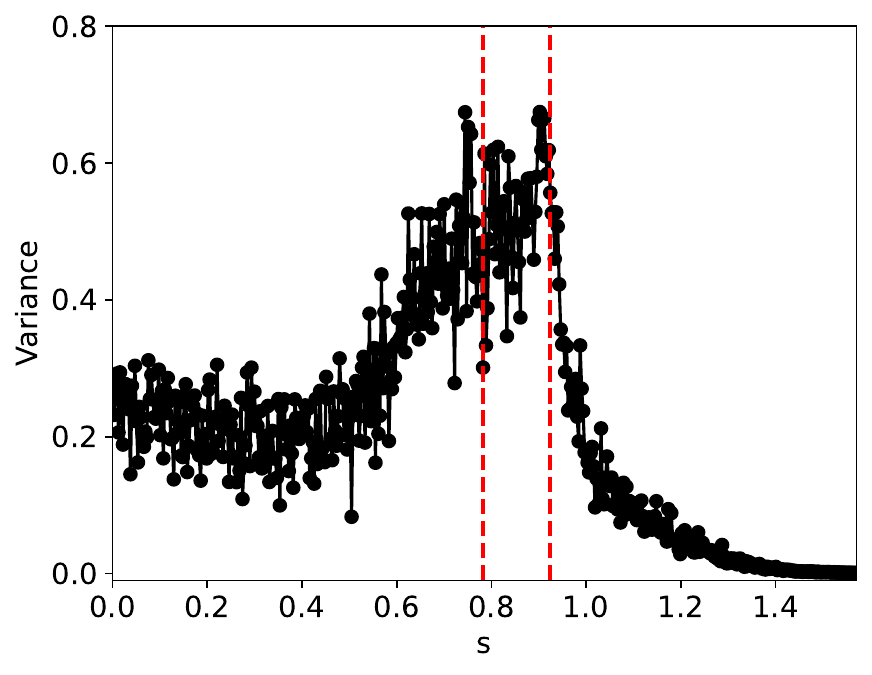}
    \end{minipage}
    \caption{Results for the SHO system at $\sigma=0.0001$, comparing MLPCL (left column) and SVDCL (right column).}
    \label{fig:SHO0.0001}
\end{figure}

\begin{figure}[t]
    \centering
    \begin{minipage}[t]{0.48\columnwidth}
        \centering
        \includegraphics[width=\linewidth]{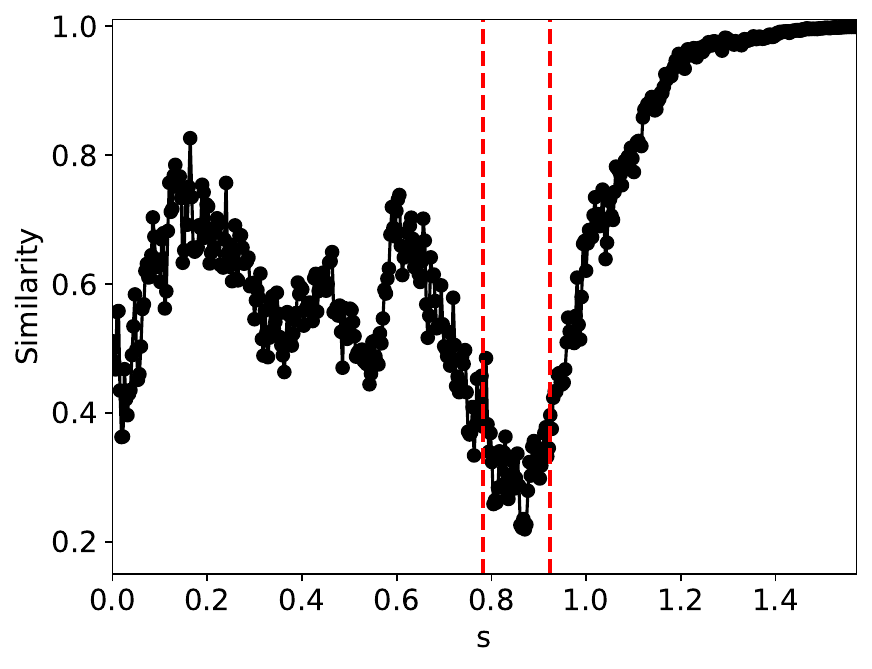}
    \end{minipage}
    \hfill
            \begin{minipage}[t]{0.48\columnwidth}
        \centering
        \includegraphics[width=\linewidth]{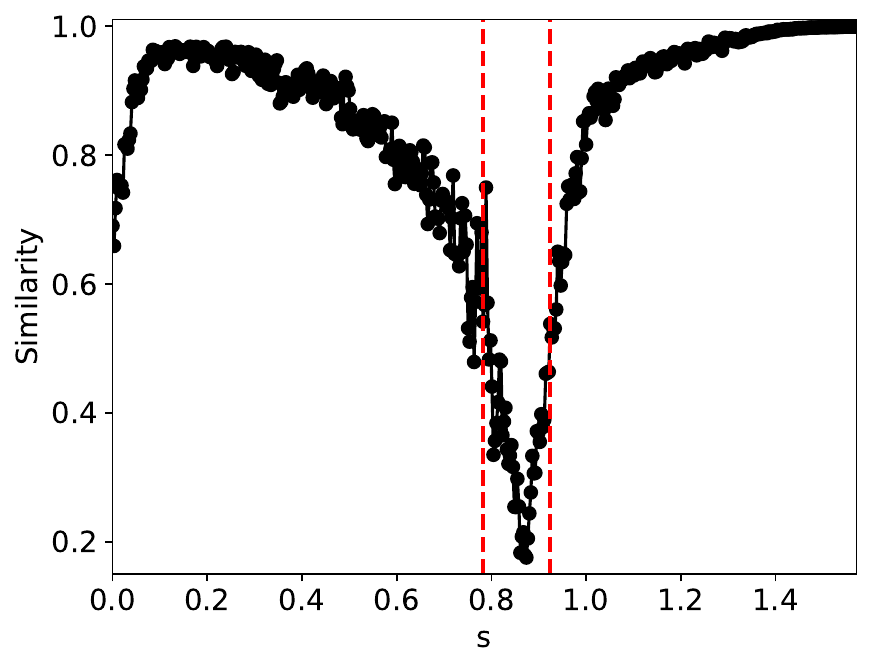}
    \end{minipage}
    \hfill
    \begin{minipage}[t]{0.48\columnwidth}
        \centering
        \includegraphics[width=\linewidth]{
        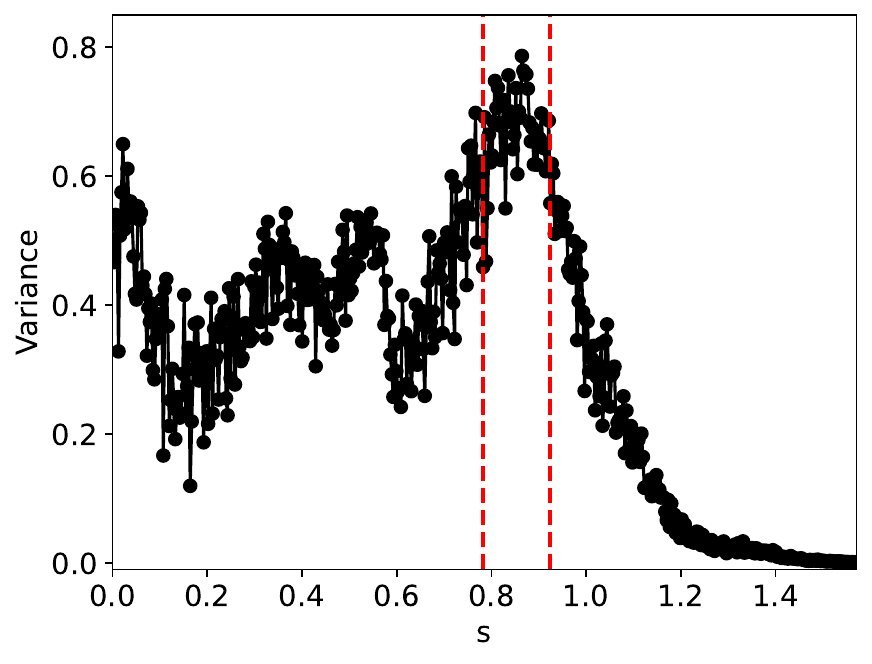}
    \end{minipage}
    \hfill
    \begin{minipage}[t]{0.48\columnwidth}
        \centering
        \includegraphics[width=\linewidth]{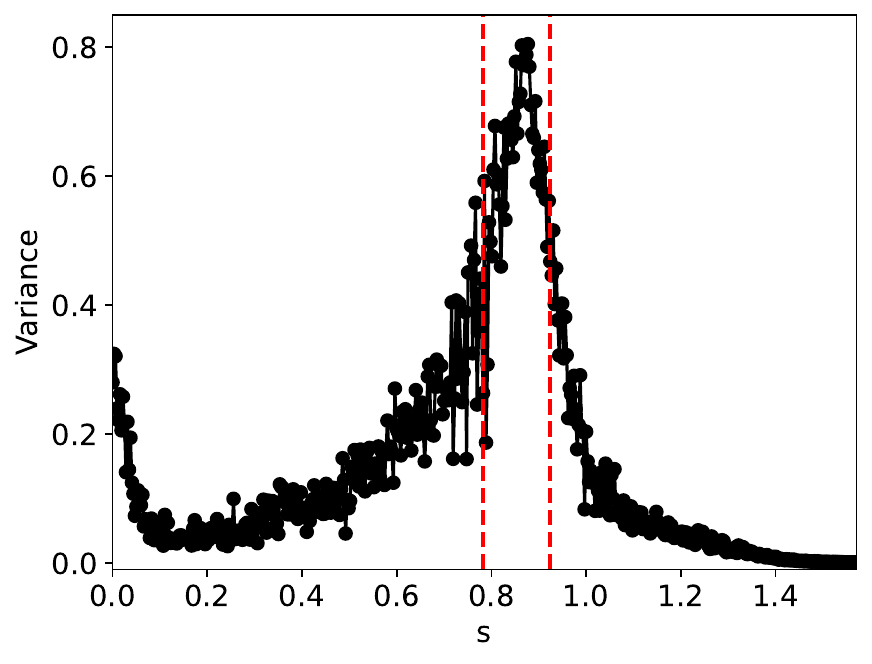}
    \end{minipage}
    \caption{The configuration matches that of Figure~\ref{fig:SHO0.0001}, with the only difference being $\sigma = 0.001$.}
    \label{fig:SHO0.001}
\end{figure}

According to the true phase diagram of the SHO system, to mitigate the effects of class imbalance \cite{abd2013review}, we oversampled the excitable and oscillatory states to approximately achieve class balance, resulting in a total of 1764 training samples (see details in Appendix~\ref{experimental setting}). The experimental results at noise levels $\sigma = 0.0001$ and $\sigma = 0.001$ are illustrated in Figures~\ref{fig:SHO0.0001} and~\ref{fig:SHO0.001}, respectively.

Under the $\sigma = 0.0001$ condition, the similarity and variance metrics exhibit two clear local extrema and correctly identify the critical transitions in the SHO system. Moreover, under this low noise level, the overall trends of the metrics remain largely unchanged from their noise-free counterparts ($\sigma = 0.0$; see Appendix~\ref{sec:classical}). Compared with MLPCL, although our method attains relatively higher similarity values near the critical points (approximately 0.3 versus 0.2 for MLPCL) and lower variance values (approximately 0.7 versus 0.8 for MLPCL), our metric curves are notably more uniform and smoother, particularly in the excitable state region. This behavior indicates superior robustness to irrelevant noise and more stable feature representation, even though our model uses approximately 20\% fewer parameters.

After increasing the noise level to $\sigma = 0.001$, Figure~\ref{fig:SHO0.001} indicates that all system states—particularly the excitable and oscillatory states—are significantly degraded, as reflected by the substantially increased spread of metric values within the same state. The metric curves of both methods approximately identify the two ground-truth critical transitions. However, MLPCL performs significantly worse, with the metrics riddled with multiple spurious local minima and maxima, making reliable detection of true critical points nearly impossible without reference information. In contrast, our method provides a more consistent and accurate indication of the true critical transitions,  reflecting its superior stability under this noisy condition.

\subsection{Non-linear cell-cycle model}
Since this model primarily captures the dynamics of cyclin, cdc2, and their complexes across different phosphorylation states, we retain only two quantities $[\mathrm{M}]$ and $[\mathrm{YT}]$, scaled by $[\mathrm{CT}]$, as the training and test dataset (see Appendix~\ref{experimental setting} for details), consistent with the notation in \cite{tyson1991modeling, romeo2025characterizing}. 
After training, the test results are presented in Figures~\ref{fig:Cellcycle0.01} and~\ref{fig:Cellcycle0.03} (see Appendix~\ref{sec:classical} for other related results).

Figure~\ref{fig:Cellcycle0.01} shows that MLPCL and SVDCL, despite having 20\% fewer parameters, yield similar patterns in both similarity and variance under the condition $\sigma = 0.01$. Close to the critical point $s_1$, both MLPCL and SVDCL undergo abrupt vertical jumps, signaling critical transitions. Around $s_2$, the metric values of SVDCL differ from MLPCL by roughly 0.2, suggesting greater distinguishability of latent features near this critical point. Additionally, the metric curves generated by SVDCL exhibit smoother and more stable behavior for large values of $s$. More intriguingly, both methods feature a sharp local extremum near $s=4$, which roughly corresponds to the boundary between the excitable and monostable states in this region, as reported in \cite{romeo2025characterizing}.

\begin{figure}[t]
    \centering
    \begin{minipage}[t]{0.48\columnwidth}
        \centering
        \includegraphics[width=\linewidth]{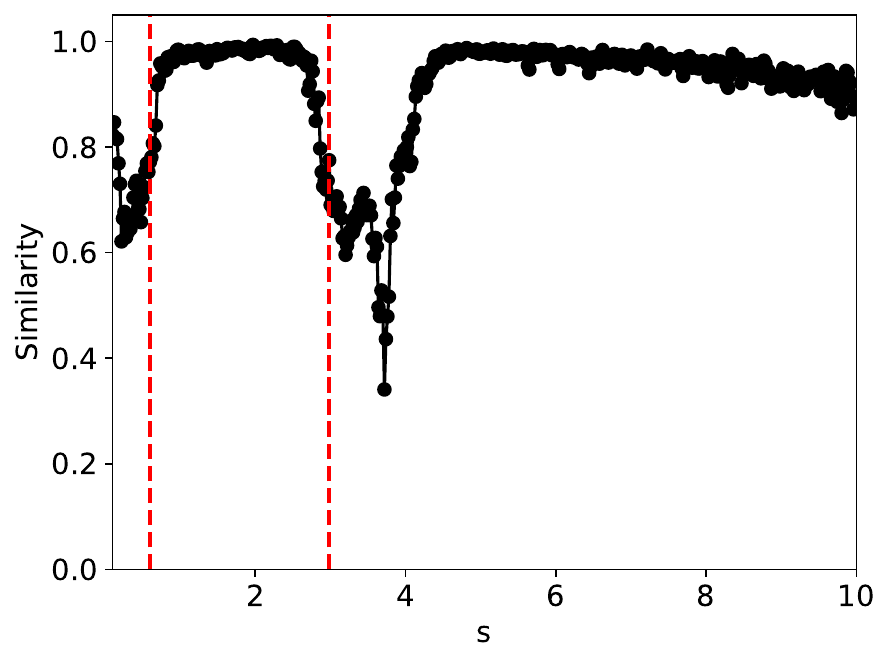}
    \end{minipage}
    \hfill
            \begin{minipage}[t]{0.48\columnwidth}
        \centering
        \includegraphics[width=\linewidth]{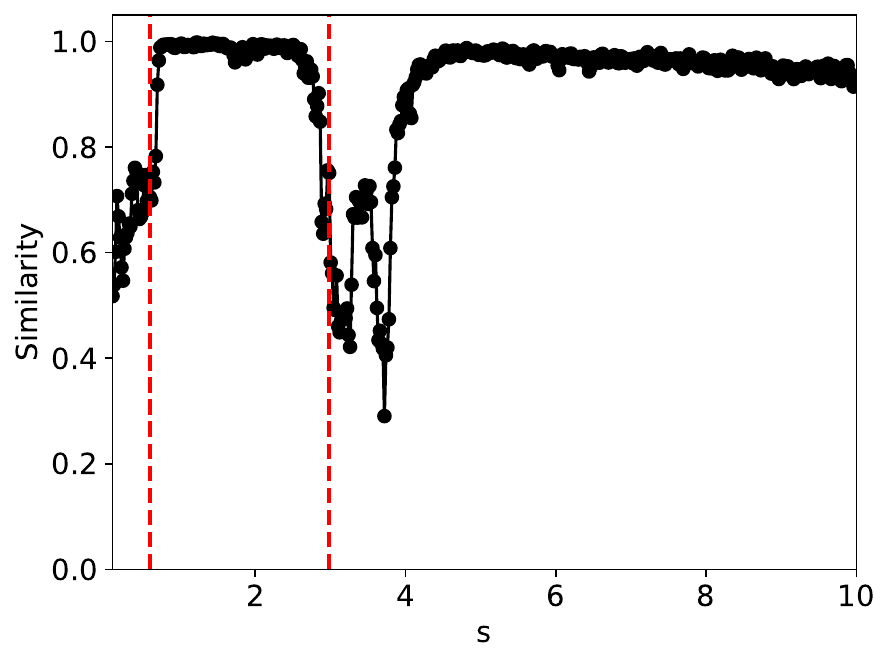}
    \end{minipage}
    \hfill
    \begin{minipage}[t]{0.48\columnwidth}
        \centering
        \includegraphics[width=\linewidth]{
        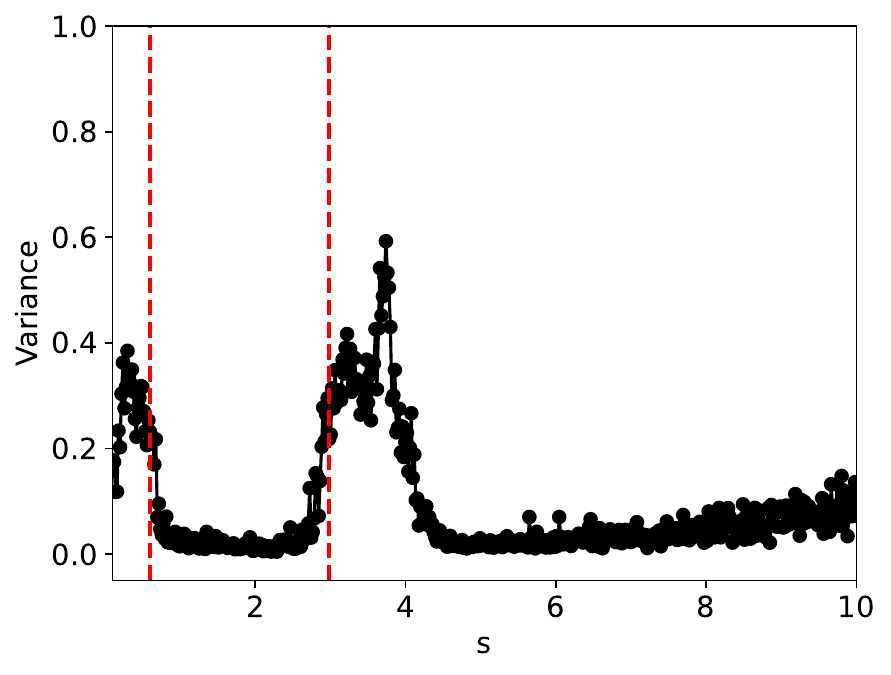}
    \end{minipage}
    \hfill
    \begin{minipage}[t]{0.48\columnwidth}
        \centering
        \includegraphics[width=\linewidth]{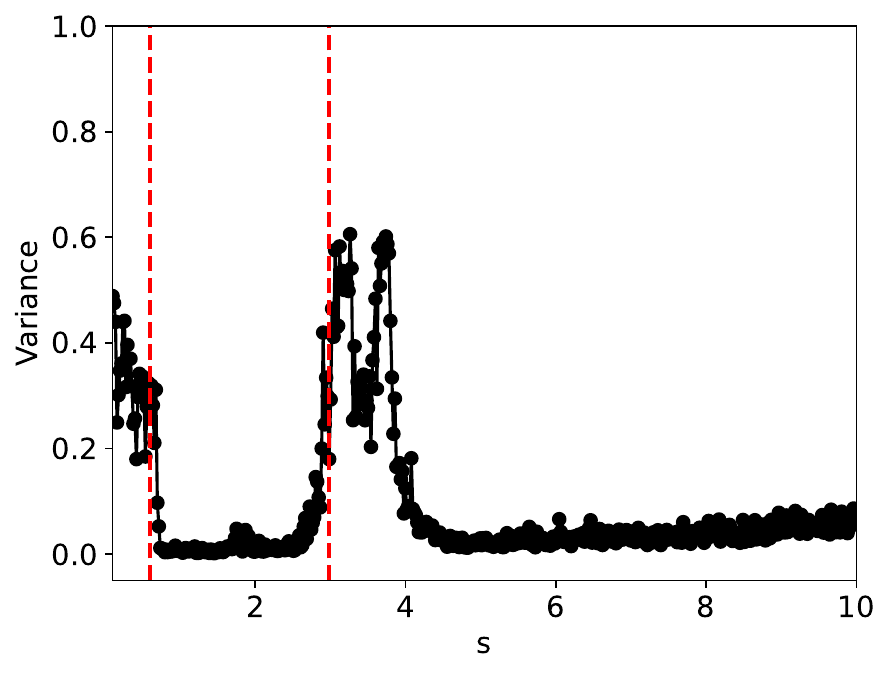}
    \end{minipage}
    \caption{Comparison result for Cellcycle system with $\sigma=0.01$: MLPCL (left column) and SVDCL (right column).}
    \label{fig:Cellcycle0.01}
\end{figure}
\begin{figure}[t]
    \centering
    \begin{minipage}[t]{0.48\columnwidth}
        \centering
        \includegraphics[width=\linewidth]{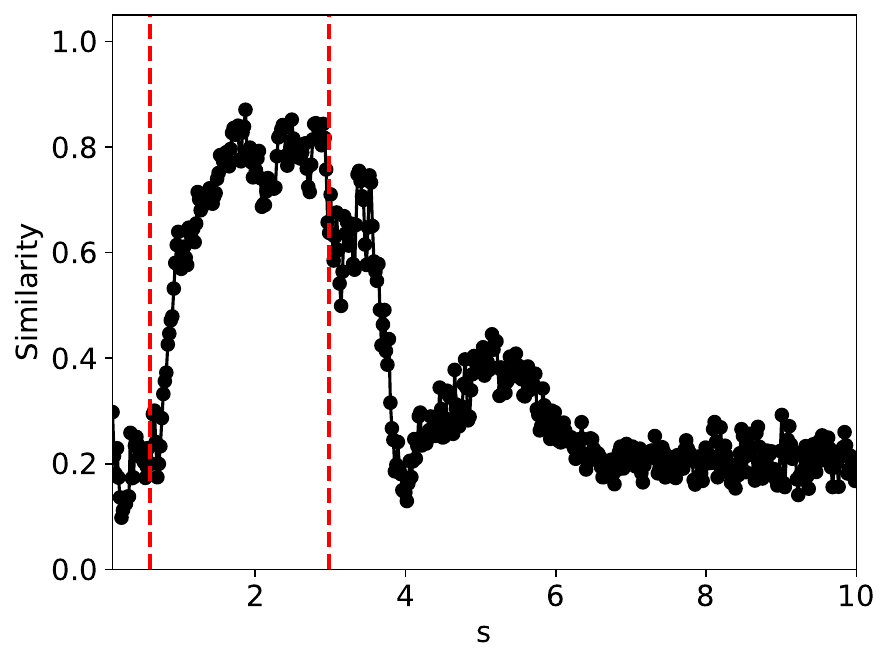}
    \end{minipage}
    \hfill
            \begin{minipage}[t]{0.48\columnwidth}
        \centering
        \includegraphics[width=\linewidth]{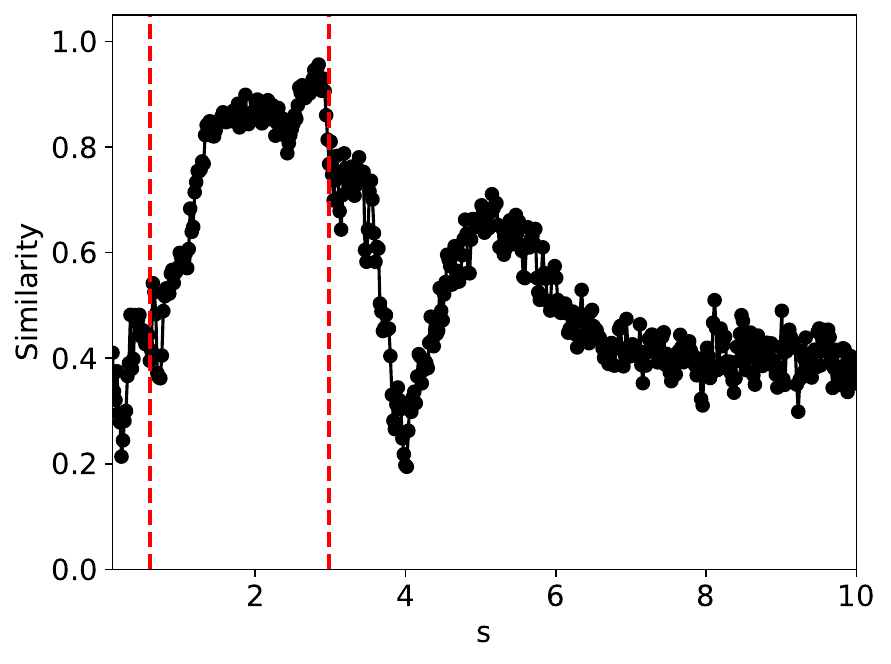}
    \end{minipage}
    \hfill
    \begin{minipage}[t]{0.48\columnwidth}
        \centering
        \includegraphics[width=\linewidth]{
        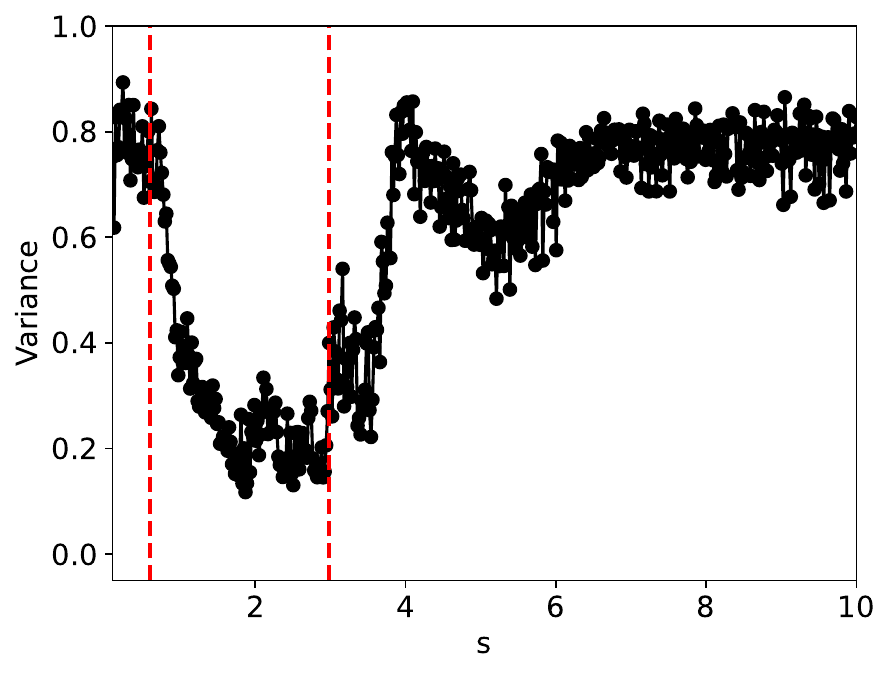}
    \end{minipage}
    \hfill
    \begin{minipage}[t]{0.48\columnwidth}
        \centering
        \includegraphics[width=\linewidth]{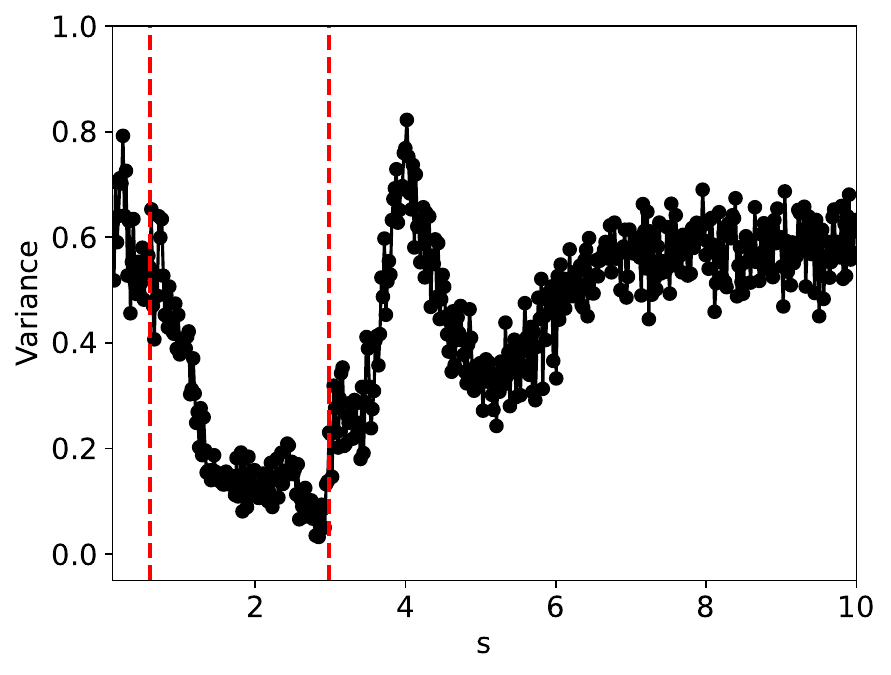}
    \end{minipage}
    \caption{Same setting as Figure \ref{fig:Cellcycle0.01}, but with $\sigma = 0.03$.}
    \label{fig:Cellcycle0.03}
\end{figure}

Besides the case of $\sigma=0.01$, we also conducted experiemnt with a higher noise level 0.03. As shown in Figure~\ref{fig:Cellcycle0.03}, the similarity and variance curves confirm that the SVDCL method is functionally equivalent to the MLPCL method, approximately capturing the full magnitude and location of the critical transitions that characterize the Cellcycle phase transitions. This demonstrates that the SVD's low-rank constraint is sufficient to detect the essential physics, making the SVDCL a highly compressed alternative to the full-rank MLPCL. Moreover, the critical point near $s=4$ appears to persist, and our method exhibits stronger structural robustness than the MLPCL approach, particularly in this region, as evidenced by the SVDCL model achieving a wider range in both similarity and variance. The larger similarity range proves that the SVDCL extracts more discriminative features, while the larger variance range confirms its greater sensitivity to underlying physical fluctuations. By leveraging its low-rank constraint as an effective spectral filter, SVDCL yields a high-fidelity, high-SNR (Signal-to-Noise Ratio) representation that is substantially less susceptible to noise corruption than MLPCL. 
Remarkably, the findings indicate the possible emergence of an additional critical point near $s=6$ under this condition, which we leave for future exploration.

\subsection{2D Ising model}
\begin{figure*}[t]
    \centering
    \begin{minipage}[t]{0.32\textwidth}
        \centering
        \includegraphics[width=\linewidth]{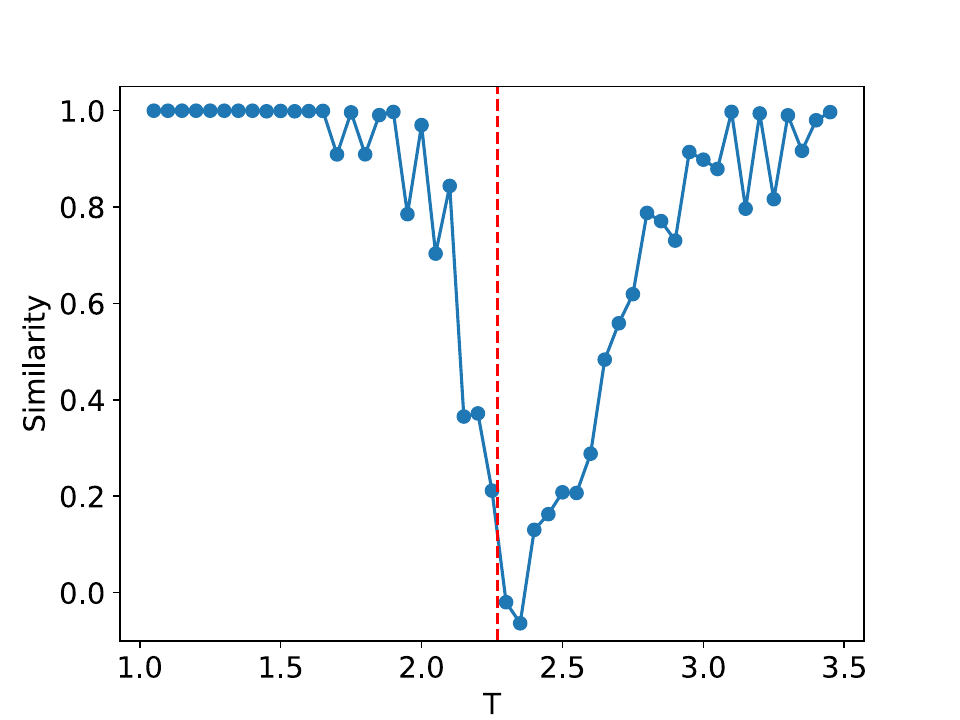}
    \end{minipage}\hfill
    \begin{minipage}[t]{0.32\textwidth}
        \centering
        \includegraphics[width=\linewidth]{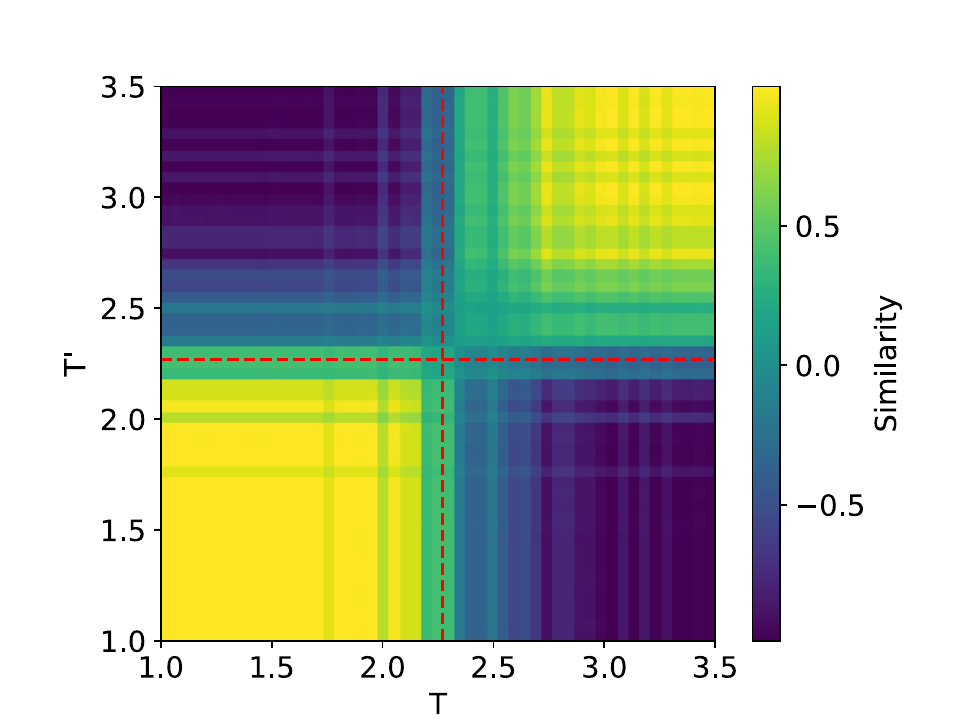}
    \end{minipage}\hfill
    \begin{minipage}[t]{0.32\textwidth}
        \centering
        \includegraphics[width=\linewidth]{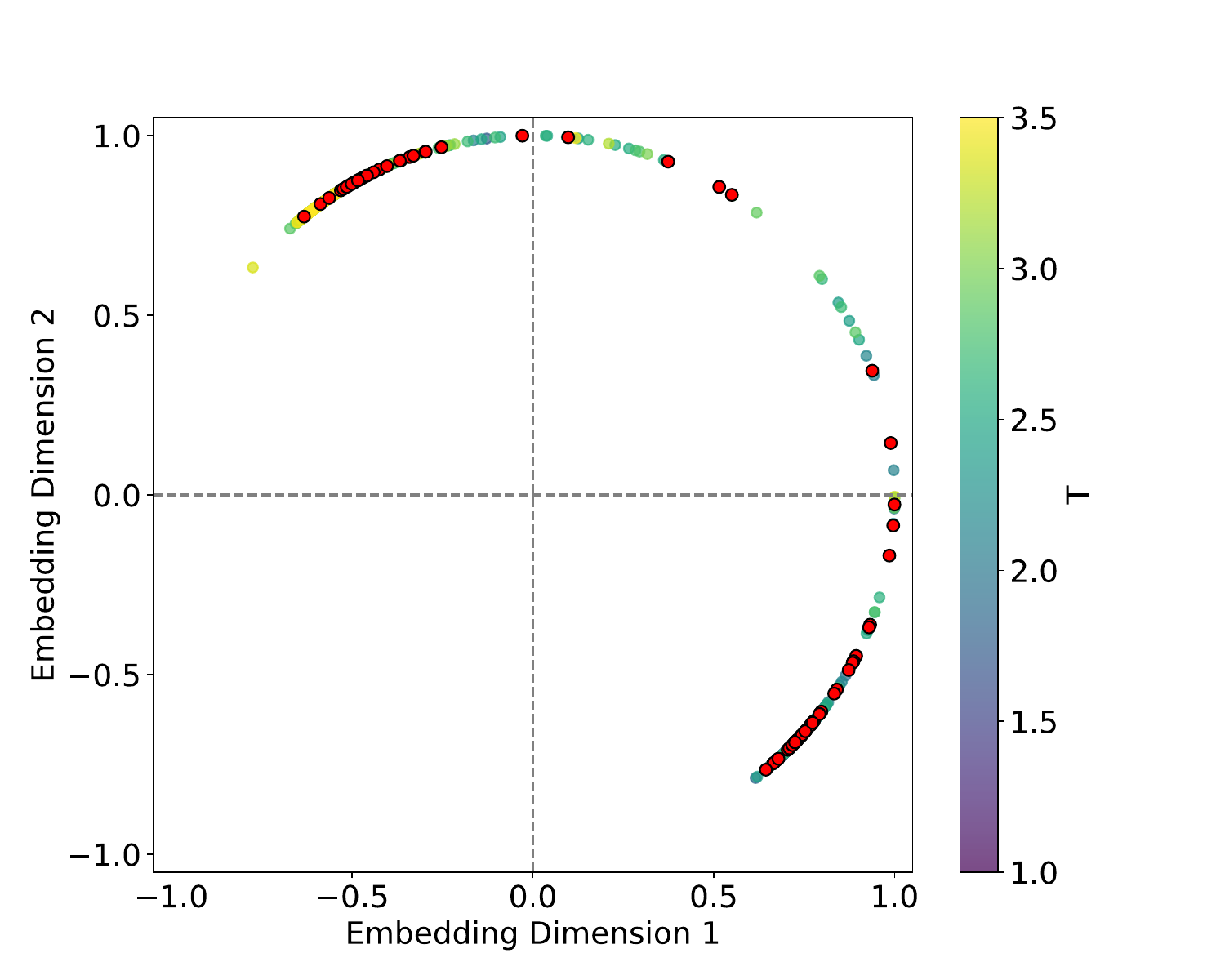}
    \end{minipage}
    \begin{minipage}[t]{0.32\textwidth}
        \centering
        \includegraphics[width=\linewidth]{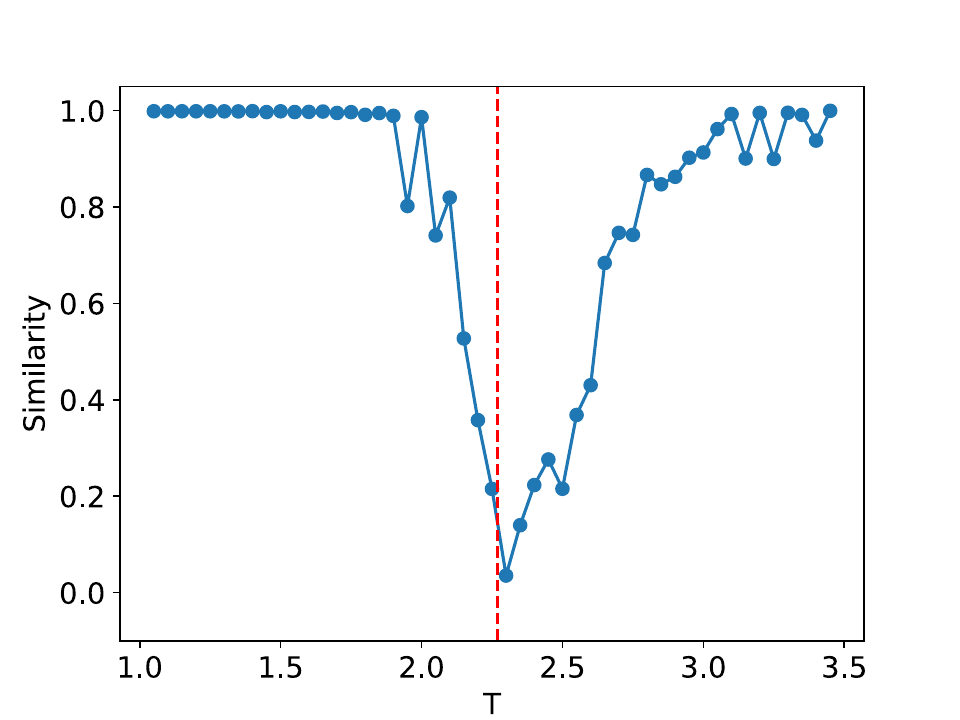}
    \end{minipage}\hfill
    \begin{minipage}[t]{0.32\textwidth}
        \centering
        \includegraphics[width=\linewidth]{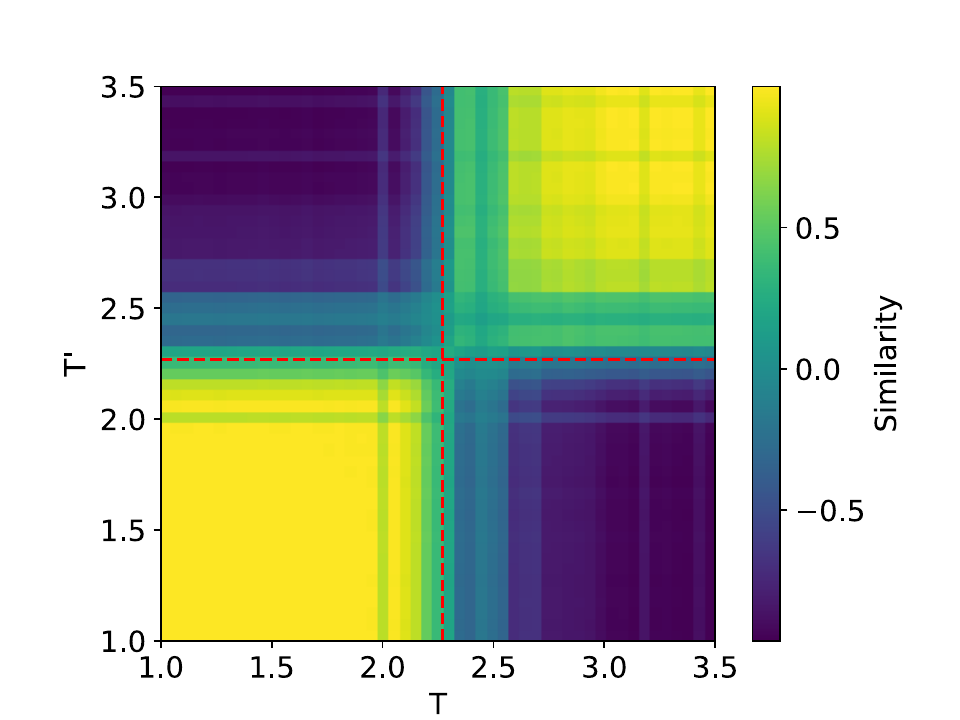}
    \end{minipage}\hfill
    \begin{minipage}[t]{0.32\textwidth}
        \centering
        \includegraphics[width=\linewidth]{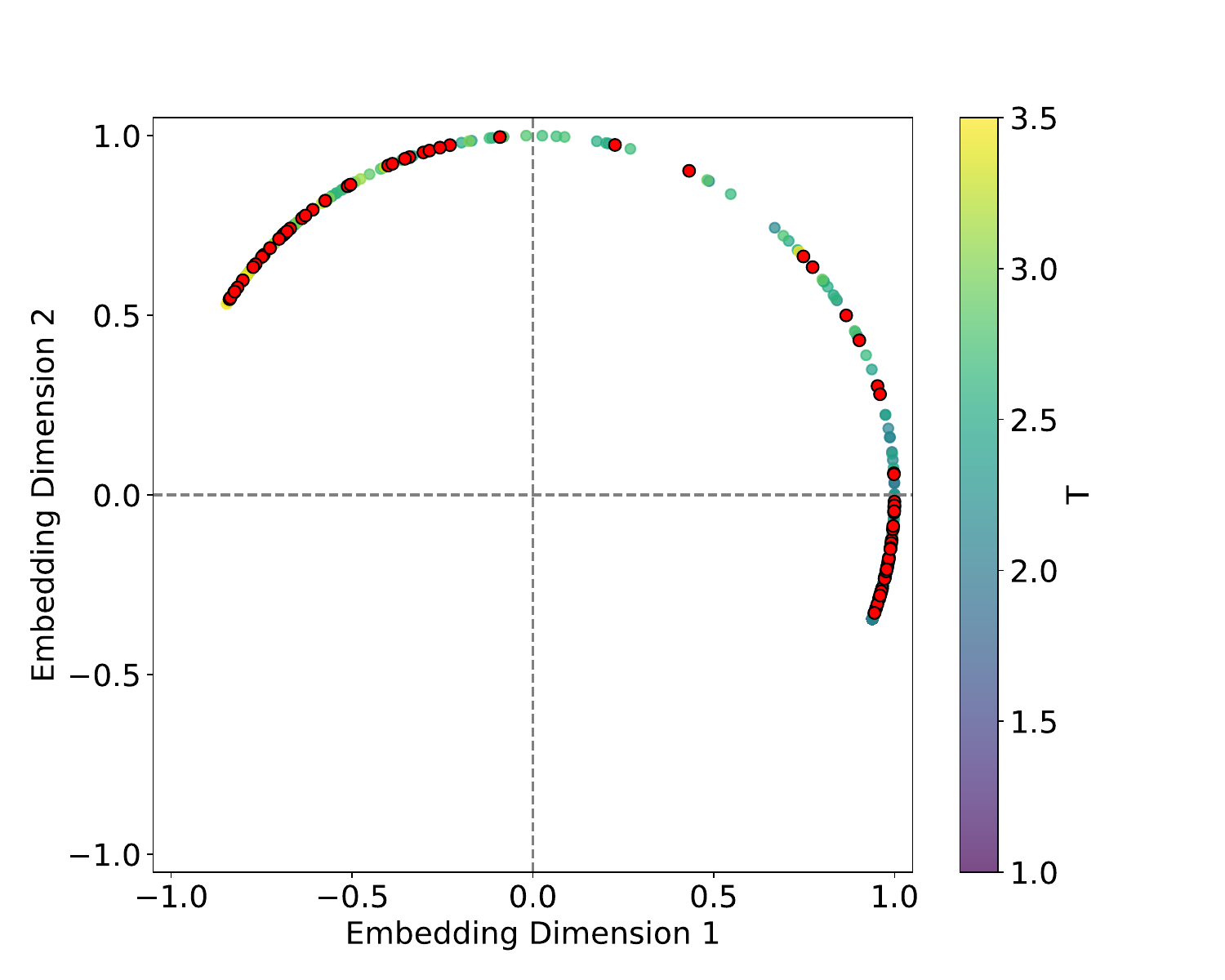}
    \end{minipage}
    \caption{Comparison results for Ising model with clean data: MLPCL (upper row) and SVDCL (bottom row).}
    \label{fig:Ising10-0.0}
\end{figure*}

For the Ising system of linear size $L$, the lattice contains $L^2$ sites. This model exhibits a well-known phase transition at the critical temperature $T_c = 2.269$, which separates the high-temperature disordered phase from the low-temperature ferromagnetic phase ~\cite{onsager1944crystal}.

The results for the Ising model ($L=10$) with noise-free data are shown in Figure~\ref{fig:Ising10-0.0} (see Appendix~\ref{experimental setting} for training settings). Despite SVDCL using only 64\% of the parameters of MLPCL, the two models yield comparable results. In both cases, the characteristic ‘V’-shaped similarity curve, computed between $T-\Delta T$ and $T+\Delta T$, faithfully reflects the true phase transition. And in areas far from the phase transition, adjacent features exhibit a high degree of similarity. the heatmap shown in the middle column of Figure \ref{fig:Ising10-0.0}, which depicts the mutual similarity between pairwise sampled $T$ and $T'$, effectively illustrates the outcome of the CL experiment and provides a more comprehensive view of the phase transition. The map is cleanly partitioned by the critical temperature $T_c$ into four distinct quadrants, confirming that both NNs successfully isolates the ordered and disordered phases. And the final normalized embeding 2D features of the spin configurations move along the circle after training. All these results agree well to the previous published result \cite{han2023framework}. 

\begin{figure}[t]
    \centering
    \begin{minipage}[t]{0.48\columnwidth}
        \centering
        \includegraphics[width=\linewidth]{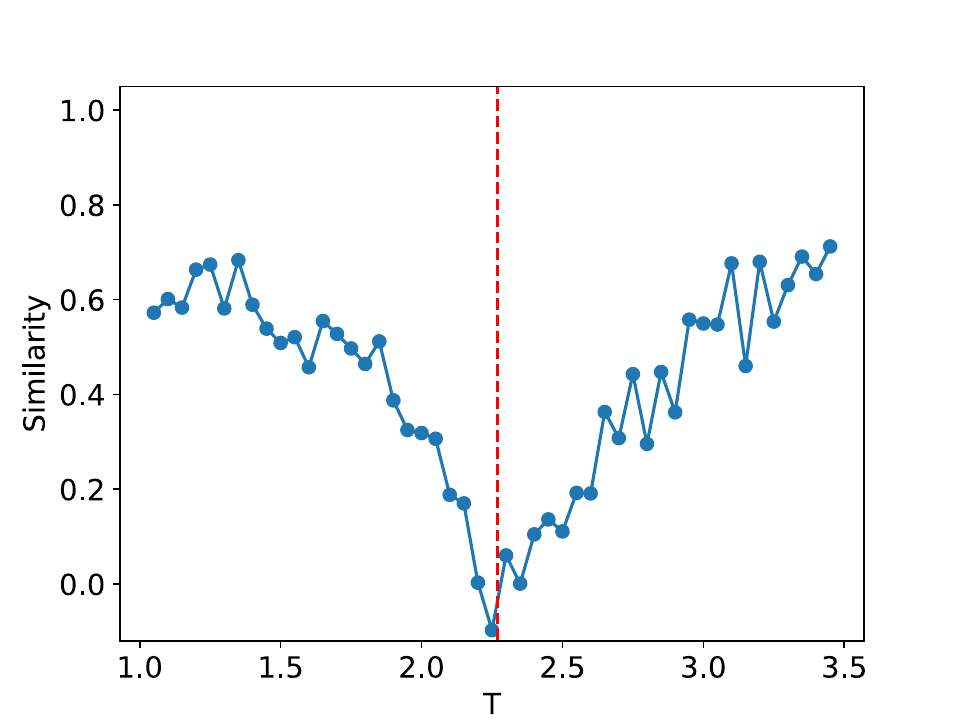}
    \end{minipage}
    \hfill
    \begin{minipage}[t]{0.48\columnwidth}
        \centering
        \includegraphics[width=\linewidth]{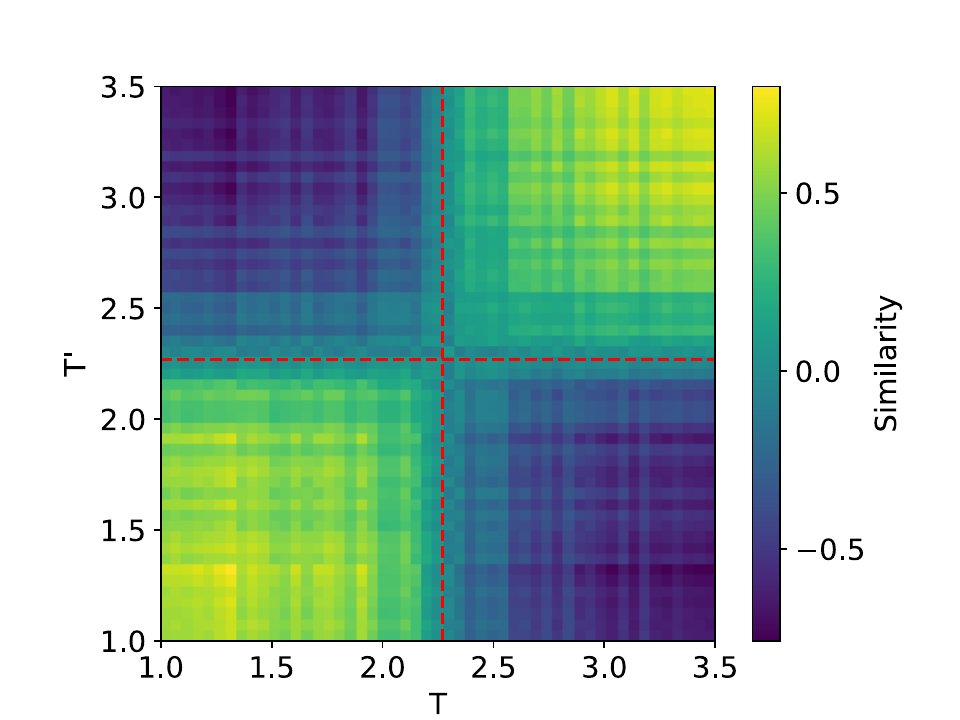}
    \end{minipage}
    \hfill
        \begin{minipage}[t]{0.48\columnwidth}
        \centering
        \includegraphics[width=\linewidth]{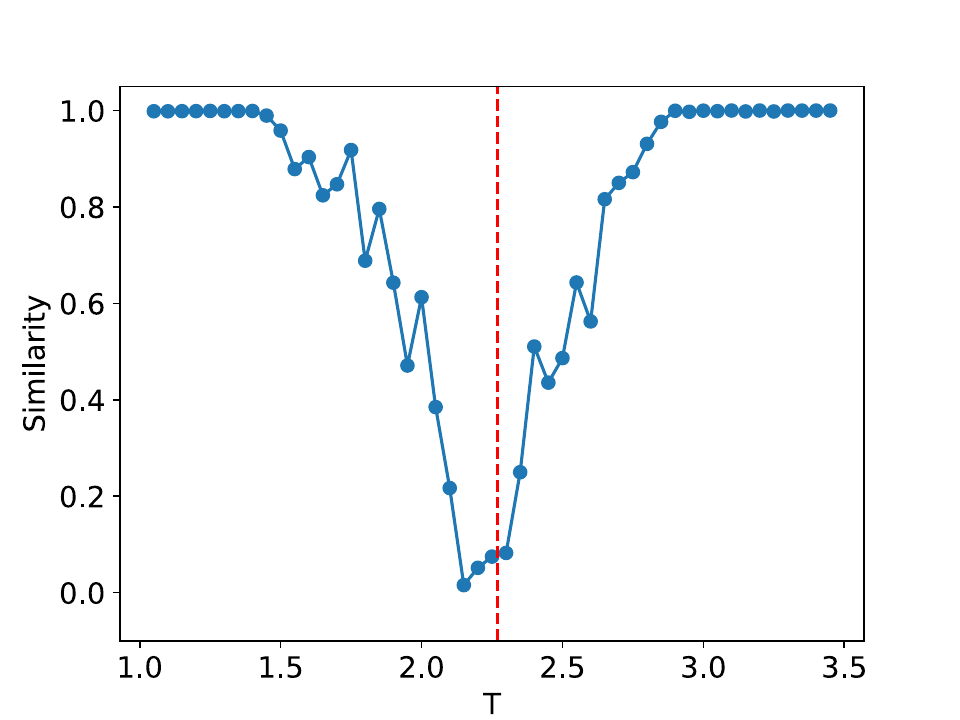}
    \end{minipage}
    \hfill
    \begin{minipage}[t]{0.48\columnwidth}
        \centering
        \includegraphics[width=\linewidth]{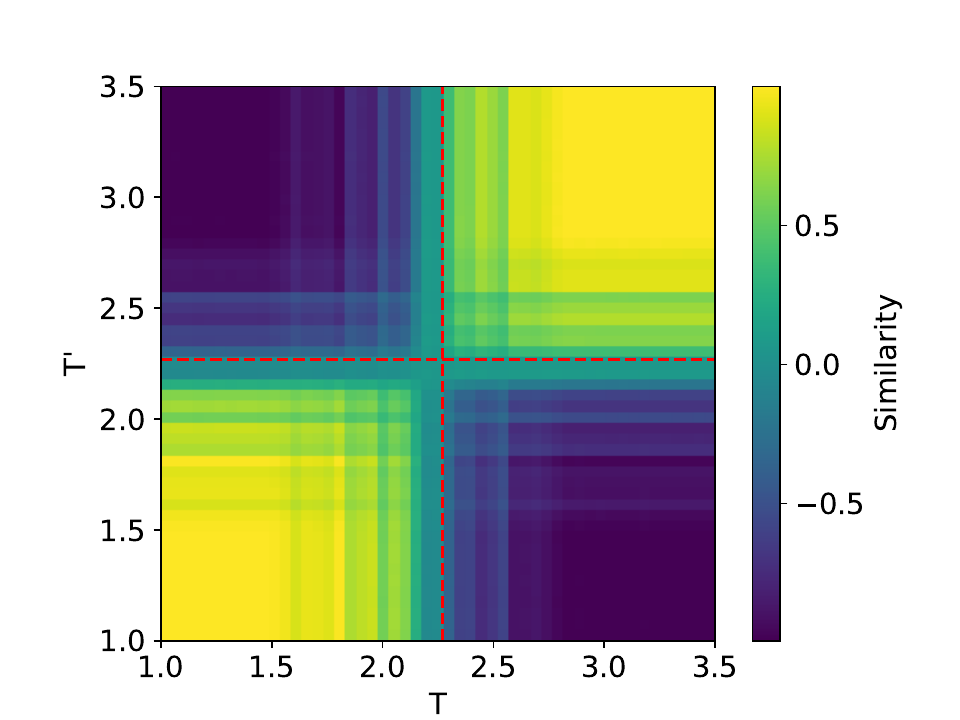}
    \end{minipage}
    \caption{Same setting as Figure~\ref{fig:Ising10-0.0}, with a noise level of 0.5: MLPCL (upper row) and SVDCL (bottom row).}
    \label{fig:Ising10-0.5}
\end{figure}

When $\sigma$ increase to 0.3, these two methods both show robust performance and get similar results to the clean data case (see Figure~\ref{fig:Ising10-0.3} in Appendix~\ref{sec:ising}). However, 
at a noise amplitude of 0.5 (Figure~\ref{fig:Ising10-0.5}), the SVDCL method remains resilient to Gaussian noise as the resulting similarity curve stays close to unity ($\approx 1.0$) throughout most of both phases and exhibits a smoother, lower-variance trend compared to MLPCL. Although the characteristic ‘V’-shape remains, the MLPCL’s jagged curve and reduced out-of-phase similarity ($\approx 0.6$) indicate that its high capacity is degraded by the noise, resulting in elevated variance and poor generalization.
And the heatmap of similarity also displays more nonuniform and irregular stripes within the same phase. In addition, we also test our model for the lattice $L=20$ case, similar patterns can be observed in our results (see Appendix~\ref{sec:ising}). These experimental results further reaffirm the robustness of our approach to noise.

\section{Discussions and conclusions}

In conclusion, we propose the SVDCL approach, which achieves strong representational power through a lightweight architecture and an elegant training strategy that rigorously enforces semi-orthogonality.
A key strength of SVDCL is its robustness to noise, arising not only from the strict semi-orthogonal constraint on the weight parameters but also from its architectural design. The SVD-based architecture naturally functions as a spectral filter, projecting representations onto a low-rank subspace while suppressing unstable or noise-dominated modes. Together with our three-stage optimization procedure that projects deviated parameters back onto the semi-orthogonal manifold, the method robustly isolates invariant structure and mitigates noise-induced overfitting. Extensive experimental results underscore the effectiveness of our method, which arises from its deliberate architectural design and principled training strategy. Compared with MLPCL, our method preserves strong capability in detecting critical transitions in dynamical systems while offering greater inference efficiency and scalability, reducing the computational cost from $\mathcal{O}(n_l n_{l+1})$ in FC layers to $\mathcal{O}(r(n_l + n_{l+1}))$ via SVD-based decomposition, and demonstrating enhanced robustness to noise. These promising results indicate that SVDCL has strong potential to advance future developments in scientific machine learning.

Nevertheless, the present study concentrates on 2D dynamical systems. 
Extending to higher dimensions substantially increases computational cost and heightens the risk of overfitting, particularly when regularization or architectural constraints are insufficient. These factors limit the model’s ability to resolve multiscale structure and to extract the latent features essential for detecting critical transitions. Moreover, as the size of the input dataset grows, training efficiency can decline due to increased memory demands and prolonged optimization. These inefficiencies may further undermine the model’s ability to represent complex scale interactions and to learn the subtle indicators of impending critical transitions. In addition, the underlying theoretical connection between critical transition phenomena and their low-rank representations remains unclear.

To overcome these limitations, future research could incorporate scalable representation-learning architectures~\cite{franceschi2019unsupervised, arel2009destin}, advances in deep causal learning~\cite{lagemann2023deep, deng2025deep}, or hierarchical graph-based neural network models~\cite{varghese2025sympgnns, yang2022hyperbolic}, thereby enhancing suitability for high-dimensional dynamical systems. Additionally,  improving the interpretability of our low-rank embedding may benefit from theoretical viewpoints such as symmetry-breaking analysis~\cite{ziyin2025parameter, ziyin2023symmetry} and geometry-aware manifold learning~\cite{meilua2024manifold, harandi2014manifold}. These perspectives can shed light on the mechanisms that make critical transitions detection possible and open new prospective pathways for advancing SVDCL and broadening its range of applications. We further hope that this work could encourage continued algorithmic and theoretical inquiry into SVDCL and the nature of critical transitions in dynamical systems.

\begin{acknowledgments}
W. Fang was funded by the National Natural Science Foundation of China (NSFC) under Grant No.12401676.
\end{acknowledgments}


\appendix

\section{Phase diagram}
\label{ground truth}

\begin{figure*}[t]
    \centering
    \begin{minipage}[t]{0.32\textwidth}
        \centering
        \includegraphics[width=\linewidth]{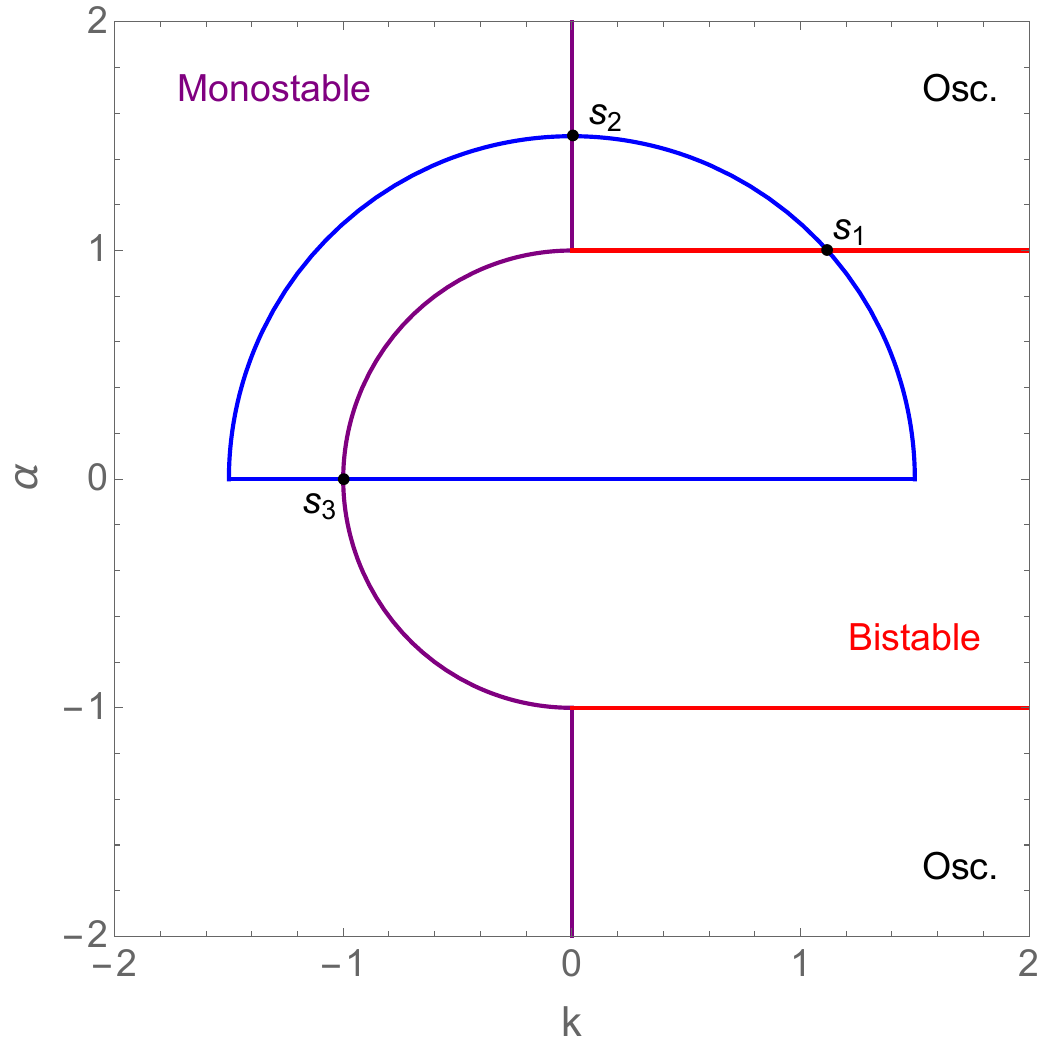}
    \end{minipage}\hfill
    \begin{minipage}[t]{0.32\textwidth}
        \centering
        \includegraphics[width=\linewidth]{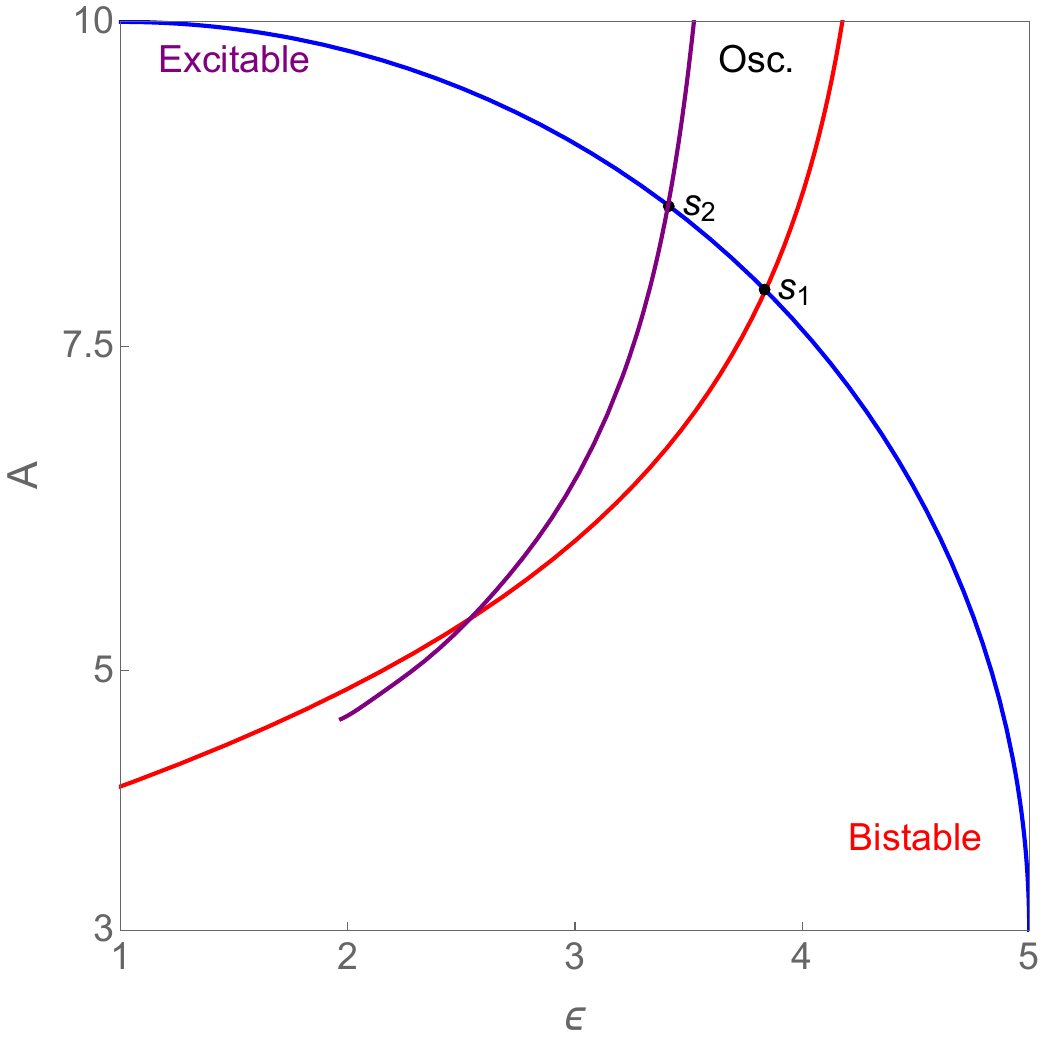}
    \end{minipage}\hfill
    \begin{minipage}[t]{0.32\textwidth}
        \centering
        \includegraphics[width=\linewidth]{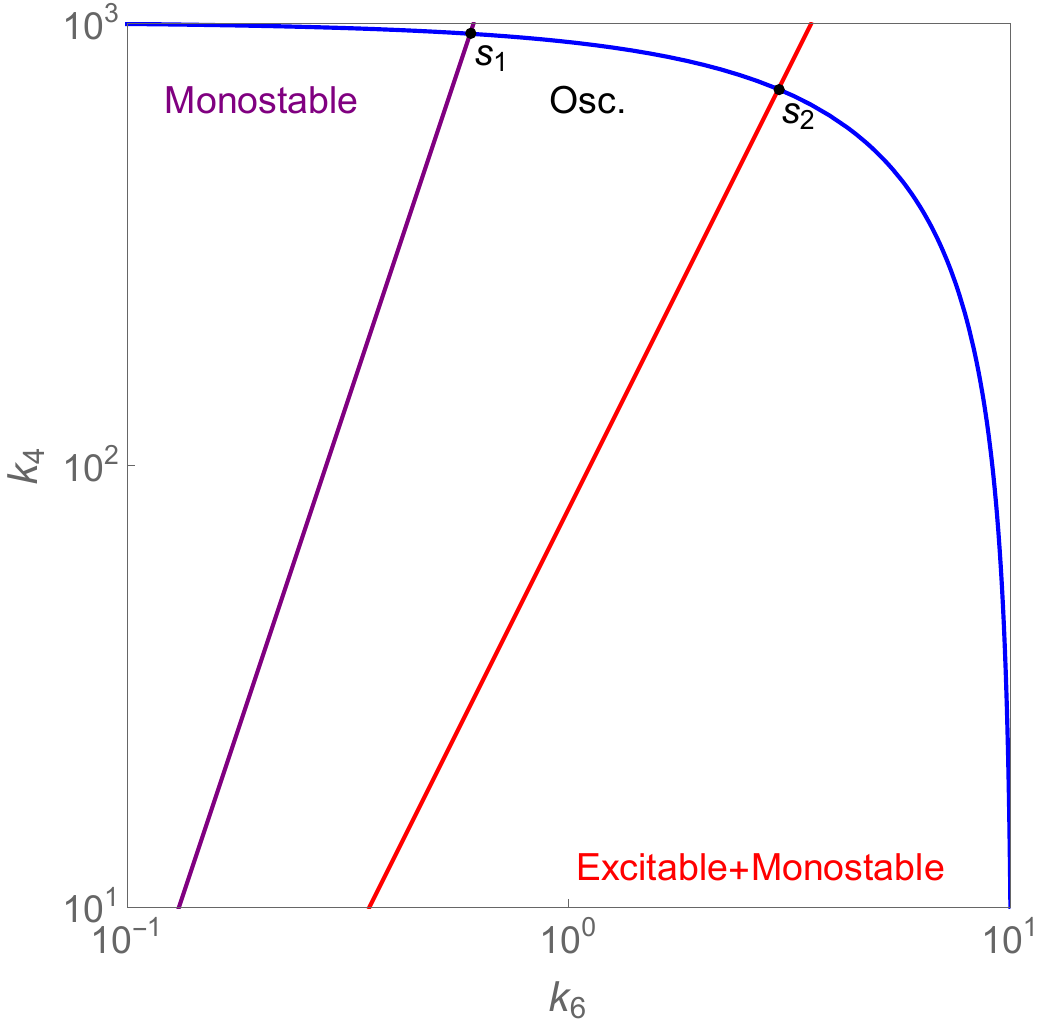}
    \end{minipage}
    \caption{Ground-truth phase diagrams for the three classical systems: (Left) SNIChopf, (Middle) SHO, and (Right) Cellcycle. The purple and red curves indicate the phase boundaries, while the blue curve denotes the selected control-parameter path used for metric evaluation. The points $s_1, s_2, s_3$ (shown as red dashed lines in the other related figures) mark the intersections between the control-parameter path and the corresponding phase boundaries.}
    \label{fig:ground truth}
\end{figure*}

Figure~\ref{fig:ground truth} provides detailed phase diagrams for the three dynamical systems. As depicted, they undergo transitions among monostable, oscillatory (Osc.), excitable, and bistable behaviors. Aside from the Cellcycle system, whose boundary between the excitable and monostable regimes varies with noise amplitude and integration timescales \cite{tyson1991modeling, romeo2025characterizing} (hence our label “Excitable+Monostable”), the remaining two systems display well-defined phase boundaries. All phase boundaries can be determined analytically, except for the purple boundary curve of the SHO system, which is computed using \texttt{Attractors.jl}~\cite{datseris2023framework} and subsequently approximated through a least-squares fitting procedure. For each system, we specify a control-parameter path that traverses all relevant dynamical regimes. From left to right in Figure~\ref{fig:ground truth}, the analytical expressions of these three control-parameter paths are:
\begin{equation}
k(s) =
\begin{cases}
1.5\cos s, & s\le\pi \\
3(s/\pi-1.5), & s>\pi
\end{cases}
, \ \ \\
\alpha(s) =
\begin{cases}
1.5\sin s, & s\le\pi\\
0, & s>\pi
\end{cases}
\end{equation}
\begin{equation}
\epsilon(s) = 4 \cos (s)+1, \ 
A(s) =7 \sin (t)+3, \ s\in [0, \pi/2]
\end{equation}
\begin{equation}
k_6(s) = s, \ k_4(s) = -100s+1010, \ s\in [0.1, 10]
\end{equation}
The intersections between the phase boundaries and the three paths, denoted $s_1, s_2, s_3$, can be obtained through straightforward analytical or numerical calculations.

For the 2D square-lattice Ising model with nearest-neighbor coupling $J$ and no external field, the exact critical temperature (marked by the red dashed line or dots) was derived by Onsager \cite{onsager1944crystal}:
\begin{equation}
T_c = \frac{2J}{k_{\mathrm B}\,\ln\!\left(1+\sqrt{2}\right)}
     \approx 2.269,
\end{equation}
where we set $J=1$ and $k_{\mathrm B}=1$ in monte carlo simulations.

\section{Experimental Setting}
\label{experimental setting}

\begin{table}
\centering
\vspace{-2mm}
\setlength{\tabcolsep}{4pt}
\begin{tabular}{lcccccc}
\toprule
\multirow{2}{*}{Case} 
& \multicolumn{3}{c}{Training data} 
& \multicolumn{3}{c}{Test data} \\
\cmidrule(lr){2-4} \cmidrule(lr){5-7}
 & number & ICs & time & number & ICs & time \\
\midrule
SNIChopf   & 1024 & 30 & 50 & 500 &  1200 & 50  \\
SHO        & 1764 & 30 & 50 & 500 &  1200 & 50 \\
Cellcycle  & 4096 & 30 & 200 & 500 & 1200 & 200  \\
Ising ($L=10$)   & 1020  & 20 & - & 1020 &  20 & - \\
Ising ($L=20$)   & 1020  & 20 & - & 1020 &  20 & - \\
\bottomrule
\end{tabular}
\vspace{-1mm}
\caption{Data setting for all the tested systems.}
\label{tab:data}
\end{table}

\textbf{Data preparation:} For the three classical systems, the training and test datasets are generated by numerically integrating the corresponding stochastic differential equations under varying random initial conditions (ICs) and control parameters, sampled at prescribed time intervals (summarized in Table~\ref{tab:data}) \cite{romeo2025characterizing, rackauckas2017differentialequations}.
The time interval was discretized into 400 points, from which 100 points were uniformly downsampled for use in both training and testing.
For the SNIChopf system, we discretize $\alpha$ and $k$ into 32 bins each, yielding $32 \times 32=1024$ training samples. Due to the markedly different sizes of the three state domains, we discretize $\epsilon$ and $A$ into 32 bins and additionally generate 490 oscillatory-state samples and 250 excitable-state samples for the SHO system. Furthermore, to remove transient effects from the early-time dynamics of the Cellcycle system, the time axis is discretized into 450 intervals, after which the first 50 samples are discarded, resulting in 400 usable time points.

For the 2D Ising model, we generate 20 spin configurations at each of 51 temperatures $T$ ($T = 1.0 + i\Delta T, i=0,...,50, \Delta T=0.05$), yielding $20 \times 51=1020$ samples in total for both training and test sets. The temperature range thus spans from 1.0 to 3.5, with all configurations generated using Monte Carlo simulations \cite{landau2021guide}.

\textbf{Data augment:} For the three classical systems, we employ the data augmentation strategy introduced in \cite{romeo2025characterizing}, which applies a random invertible linear transformation based on SVD decomposition. For the Ising model, data augmentation was performed using random spin flips (up–down and grid exchange) and rotations by $\frac{n\pi}{2}$ for $n = 0, 1, 2, 3$.

\textbf{Neural network settings:} The detailed network architectures of MLPCL and SVDCL are summarized in Table~\ref{tab:nn}. For the three classical systems, the networks consist of three layers, whereas two layers are used for the Ising model. The layer widths $n_l$ and ranks $r$ are specified accordingly. The total number of parameters for both MLPCL and SVDCL is also reported. Bias terms are excluded in all experiments with our method, except for those involving the Ising model.

\begin{table}
\centering
\vspace{-2mm}
\setlength{\tabcolsep}{4pt}
\begin{tabular}{lcccc}
\toprule
\multirow{1}{*}{Case} 
& \multicolumn{1}{c}{Layer 1} 
& \multicolumn{1}{c}{Layer 2}
& \multicolumn{1}{c}{Layer 3} 
& \multicolumn{1}{c}{Parameter count}\\
\midrule
SNIChopf   & 128/100 & 128/50 & 128/50 &   801152/638600 \\
SHO        & 128/100 & 128/50 & 128/50 &   801152/638600 \\
Cellcycle  & 128/100 & 128/50 & 128/50 & 801152/638600 \\ 
Ising ($L=10$)  & 5/3  & 2/1 & - &  517/333 \\
Ising ($L=20$)  & 5/3  & 2/1 & - &  2017/1233 \\
\bottomrule
\end{tabular}
\vspace{-1mm}
\caption{Neural network configurations for MLPCL and SVDCL ($n_l/r$).}
\label{tab:nn}
\end{table}

\section{Additional results for three classical systems}
\label{sec:classical}
\begin{figure}[t]
    \centering
    \begin{minipage}[t]{0.48\columnwidth}
        \centering
        \includegraphics[width=\linewidth]{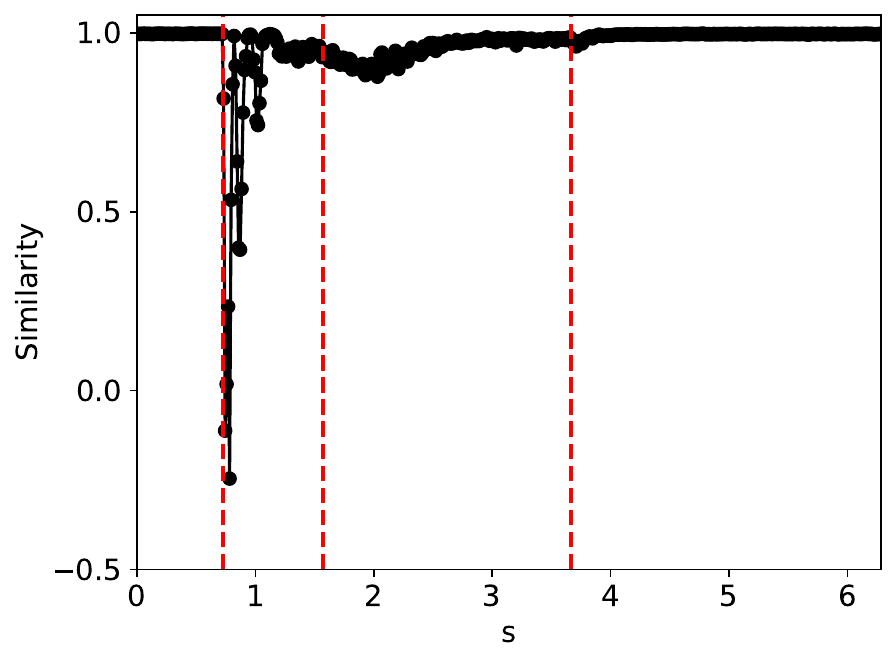}
    \end{minipage}
    \hfill
            \begin{minipage}[t]{0.48\columnwidth}
        \centering
        \includegraphics[width=\linewidth]{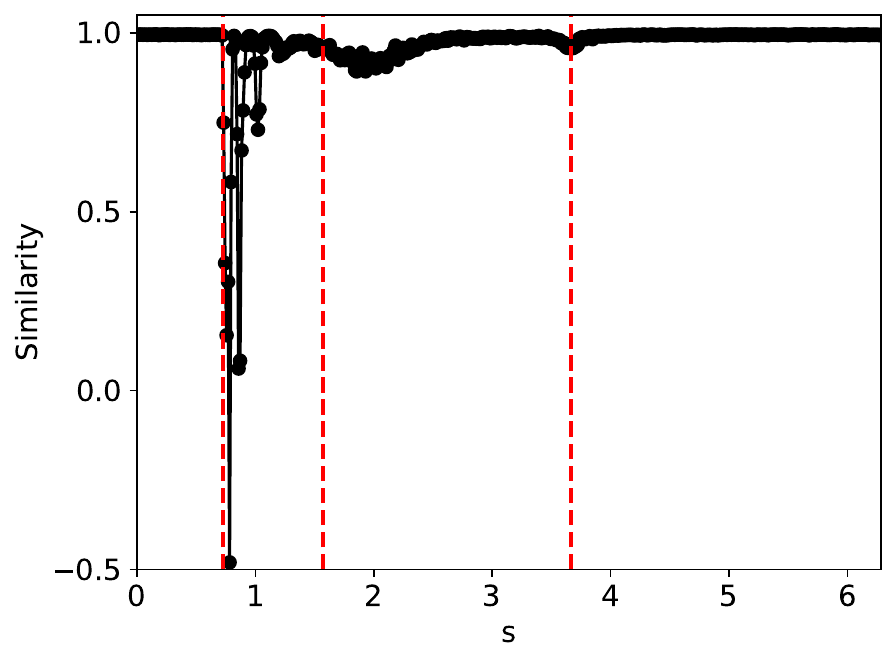}
    \end{minipage}
    \hfill
    \begin{minipage}[t]{0.48\columnwidth}
        \centering
        \includegraphics[width=\linewidth]{
        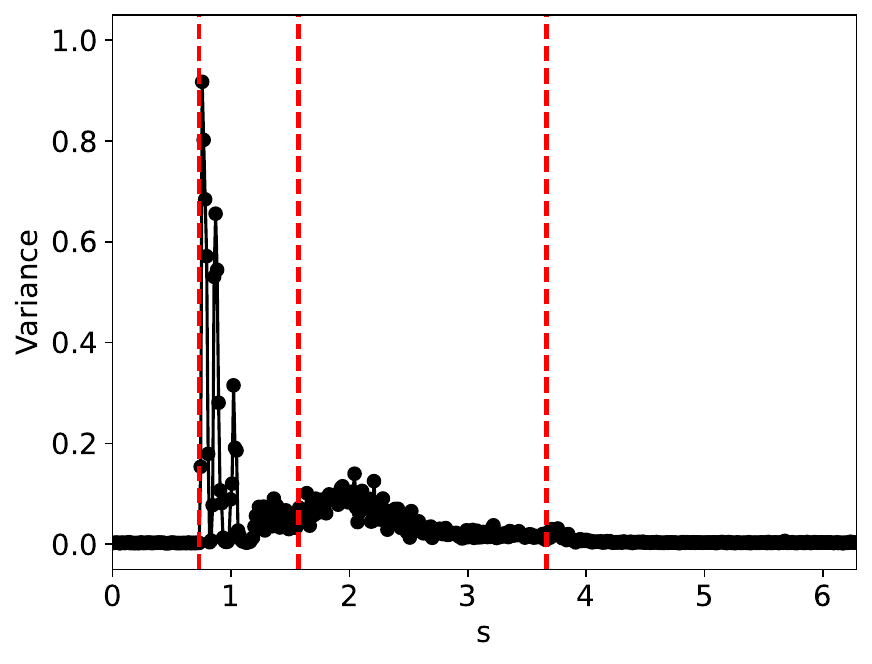}
    \end{minipage}
    \hfill
    \begin{minipage}[t]{0.48\columnwidth}
        \centering
        \includegraphics[width=\linewidth]{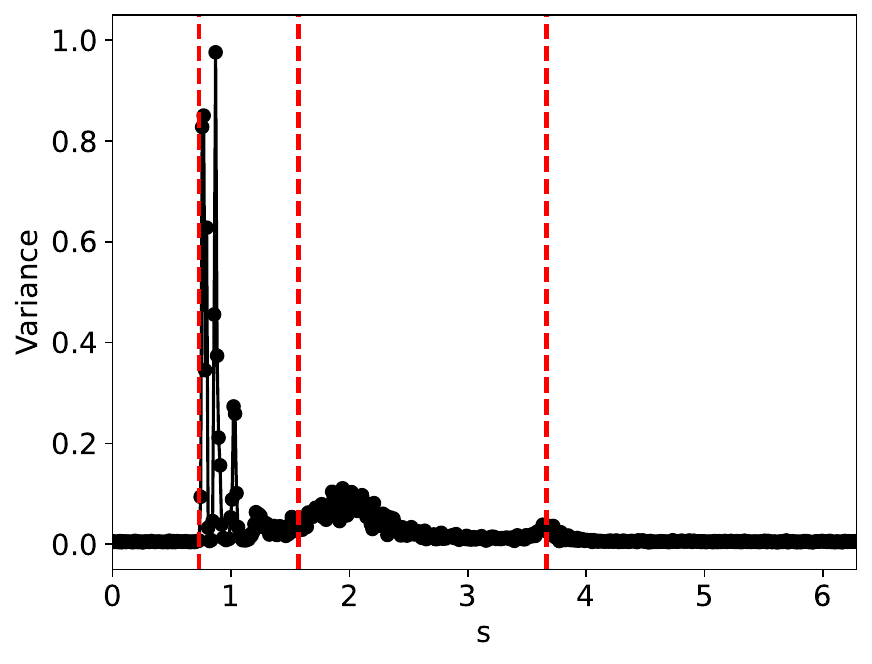}
    \end{minipage}
    \caption{The setting is identical to that of Figure~\ref{fig:SNIChopf0.005}, except with $\sigma = 0.0$.}
    \label{fig:SNIChopf0.0}
\end{figure}

\textbf{SNIChopf:} We conducted experiments with $\sigma = 0.0$, and the results are shown in Figure~\ref{fig:SNIChopf0.0}. MLPCL and SVDCL exhibit similar behavior, characterized by large peaks and oscillations around the critical point near $s_1$, as well as two smaller local bumps in the vicinity of $s_2$ and $s_3$. This indicates that the proposed SVDCL approach can achieve performance comparable to that of MLPCL while reducing the number of parameters by approximately 20\% with clean data.

\begin{figure}[t]
    \centering
    \begin{minipage}[t]{0.48\columnwidth}
        \centering
        \includegraphics[width=\linewidth]{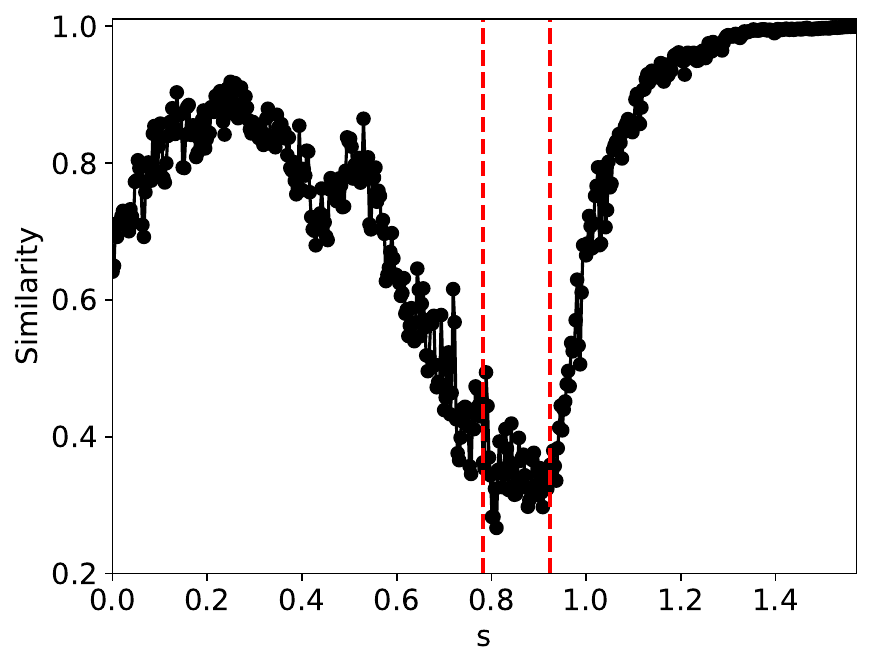}
    \end{minipage}
    \hfill
            \begin{minipage}[t]{0.48\columnwidth}
        \centering
        \includegraphics[width=\linewidth]{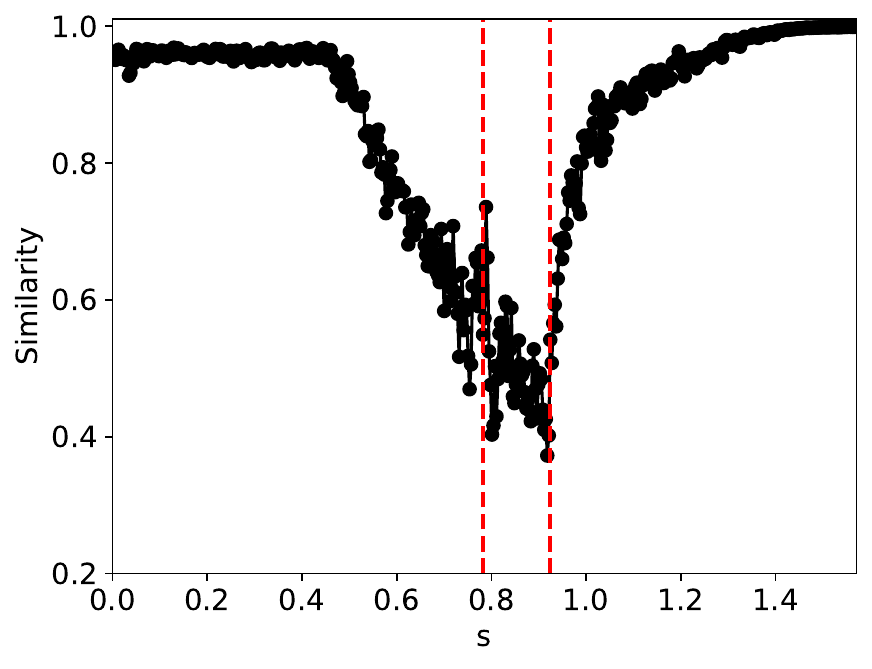}
    \end{minipage}
    \hfill
    \begin{minipage}[t]{0.48\columnwidth}
        \centering
        \includegraphics[width=\linewidth]{
        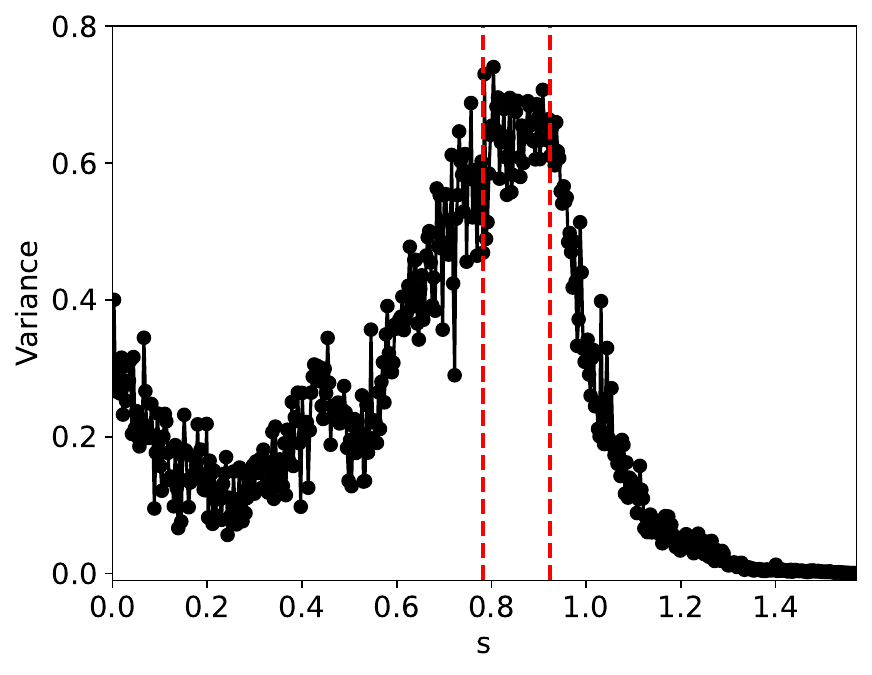}
    \end{minipage}
    \hfill
    \begin{minipage}[t]{0.48\columnwidth}
        \centering
        \includegraphics[width=\linewidth]{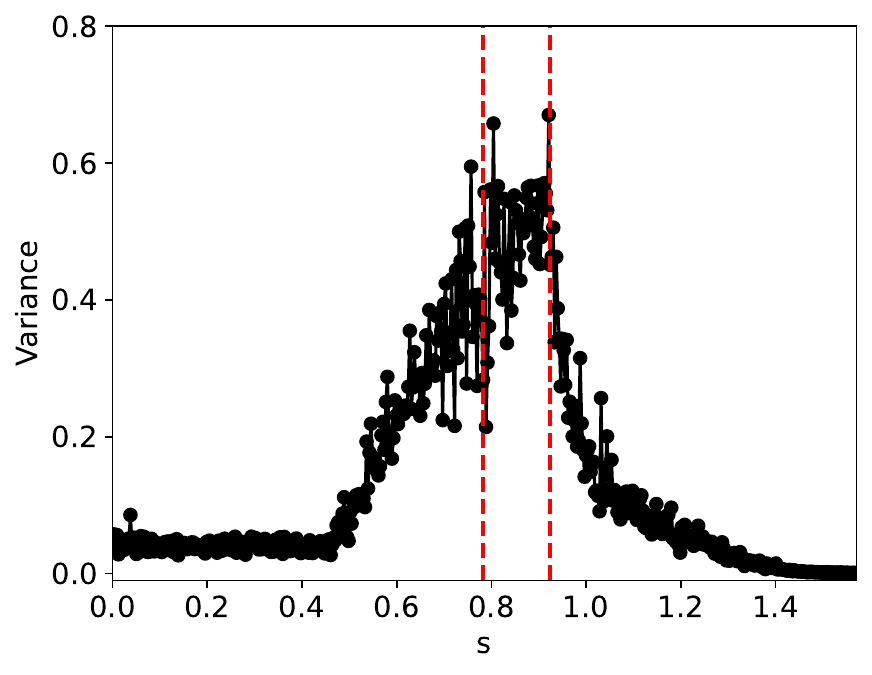}
    \end{minipage}
    \caption{Same setting as Figure~\ref{fig:SHO0.0001}, but with $\sigma = 0.0$.}
    \label{fig:SHO0.0}
\end{figure}

\textbf{SHO:} To further evaluate our method, an additional experiment was conducted using clean training data. The results are shown in Figure~\ref{fig:SHO0.0}. These two methods accurately detect the critical transition and show good agreement with the ground truth. Although MLPCL achieves slightly better metric values around critical points, it also exhibits an additional small local extremum away from the true critical points, which may create the false impression of another critical transition. Moreover, SVDCL produces more uniform and better metric values in the excitable-state region compared with MLPCL.

\textbf{Cellcycle:} 
\begin{figure}[t]
    \centering
    \begin{minipage}[t]{0.48\columnwidth}
        \centering
        \includegraphics[width=\linewidth]{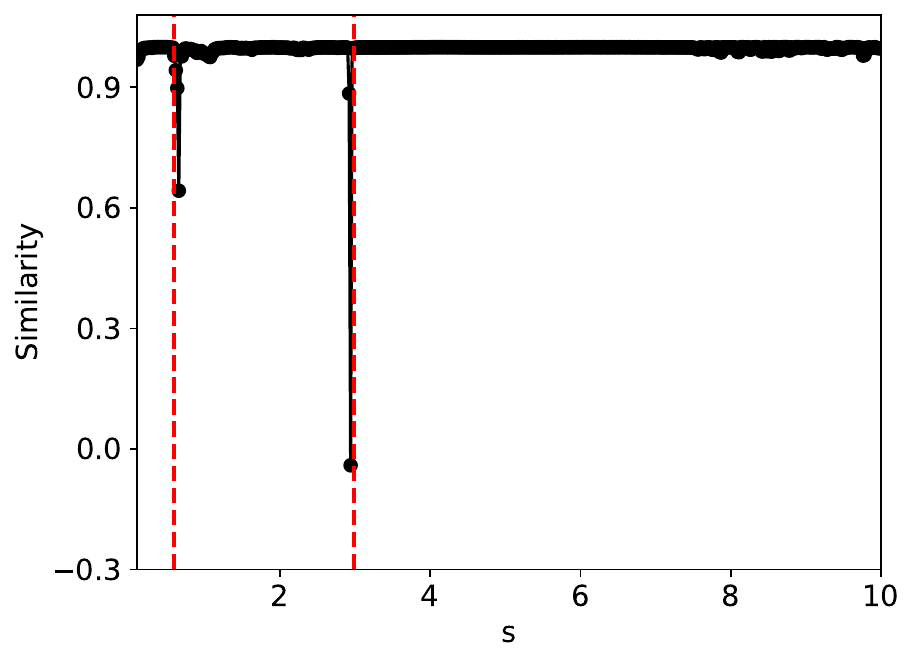}
    \end{minipage}
    \hfill
            \begin{minipage}[t]{0.48\columnwidth}
        \centering
        \includegraphics[width=\linewidth]{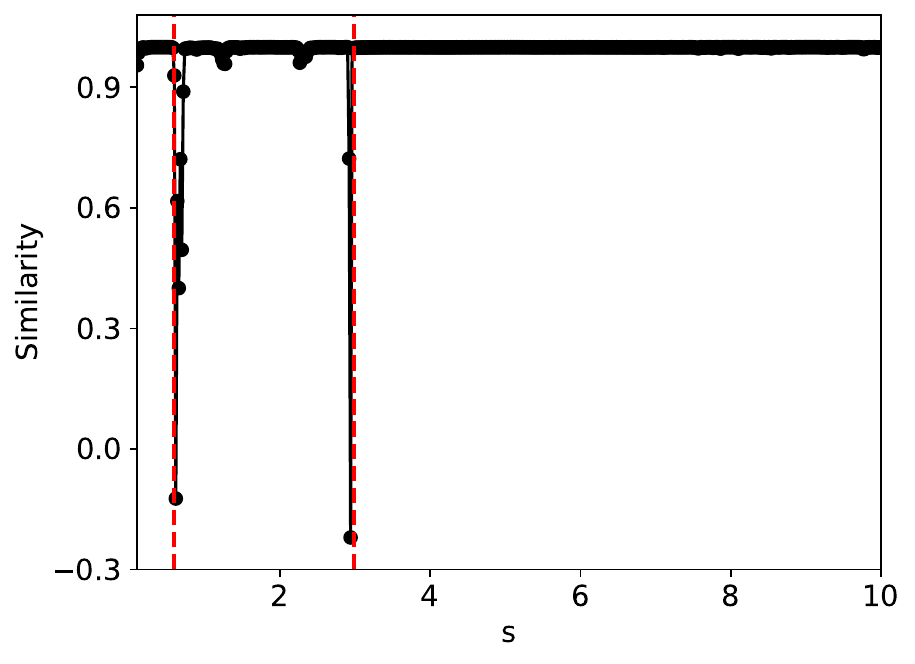}
    \end{minipage}
    \hfill
    \begin{minipage}[t]{0.48\columnwidth}
        \centering
        \includegraphics[width=\linewidth]{
        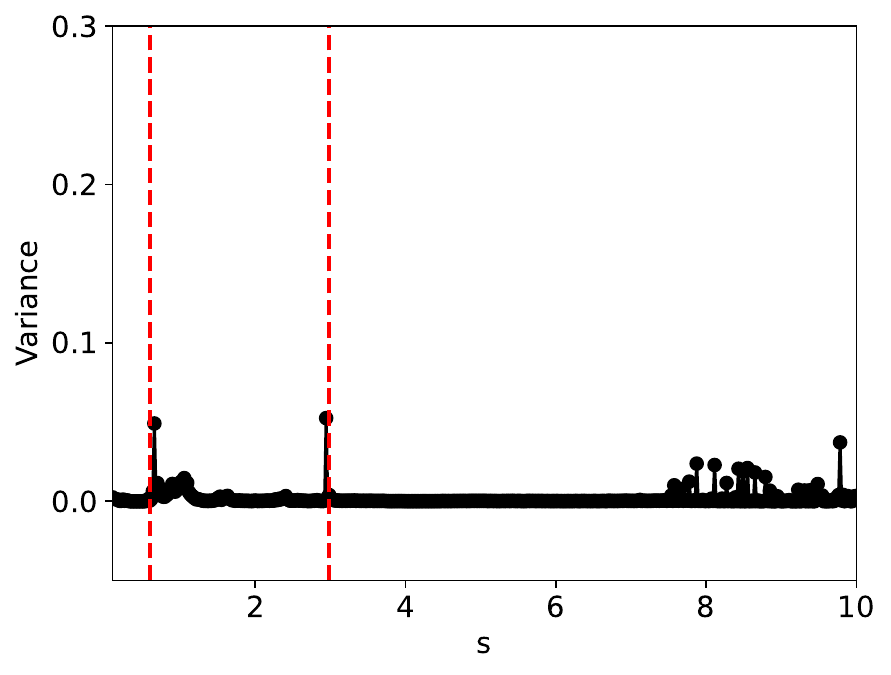}
    \end{minipage}
    \hfill
    \begin{minipage}[t]{0.48\columnwidth}
        \centering
        \includegraphics[width=\linewidth]{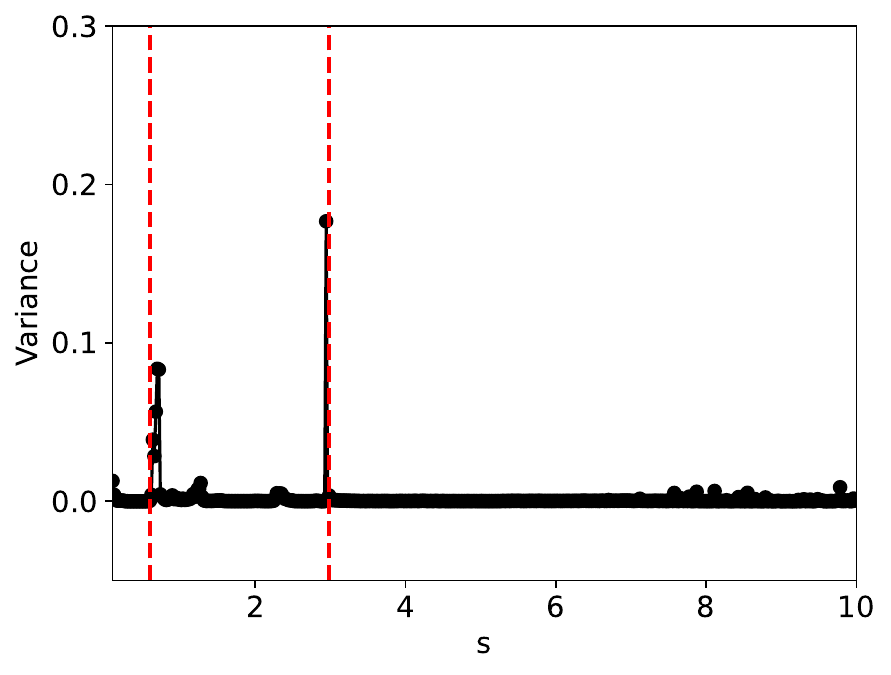}
    \end{minipage}
    \caption{The setting is identical to that of Figure \ref{fig:Cellcycle0.01}, except with $\sigma = 0.0$.}
    \label{fig:Cellcycle0.0}
\end{figure}
For the noise-free case, the results 
are illustrated in Figure~\ref{fig:Cellcycle0.0}. The two ground-truth critical transitions are accurately captured, and our SVDCL exhibits much more conspicuous indicator values than MLPCL. In addition, the variance produced by the MLPCL method exhibits several irregular local maxima when $s \ge 8$, while the other three subfigures suggest that no further critical transitions are present. This discrepancy may stem from poor training of the over-parameterized neural network in MLPCL, thereby influencing the final inference performance.
\begin{figure}[t]
    \centering
    \begin{minipage}[t]{0.48\columnwidth}
        \centering
        \includegraphics[width=\linewidth]{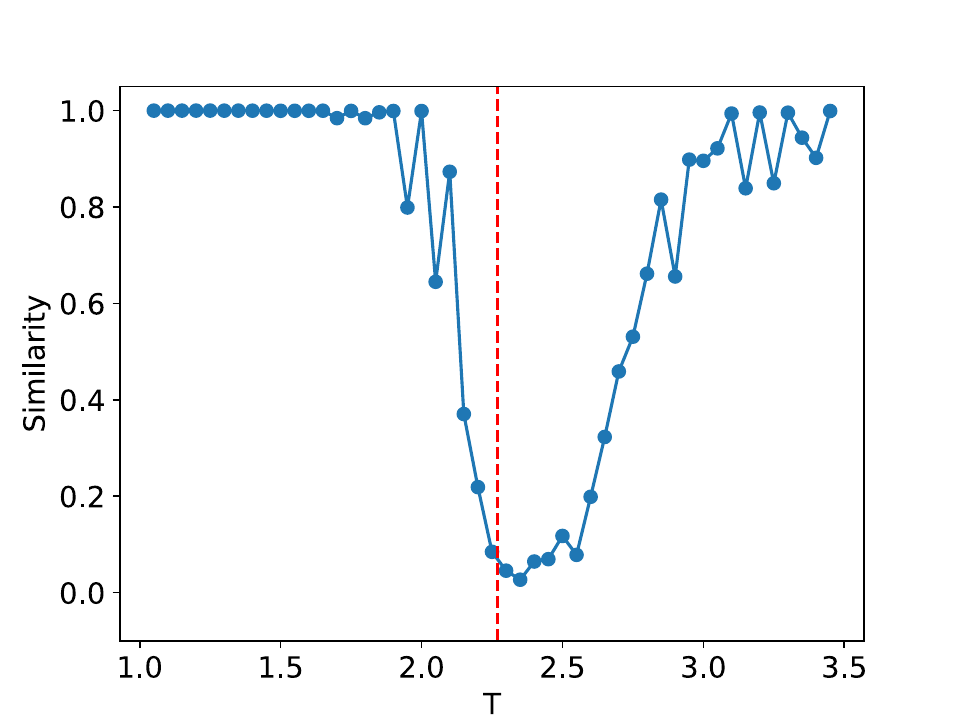}
    \end{minipage}
    \hfill
    \begin{minipage}[t]{0.48\columnwidth}
        \centering
        \includegraphics[width=\linewidth]{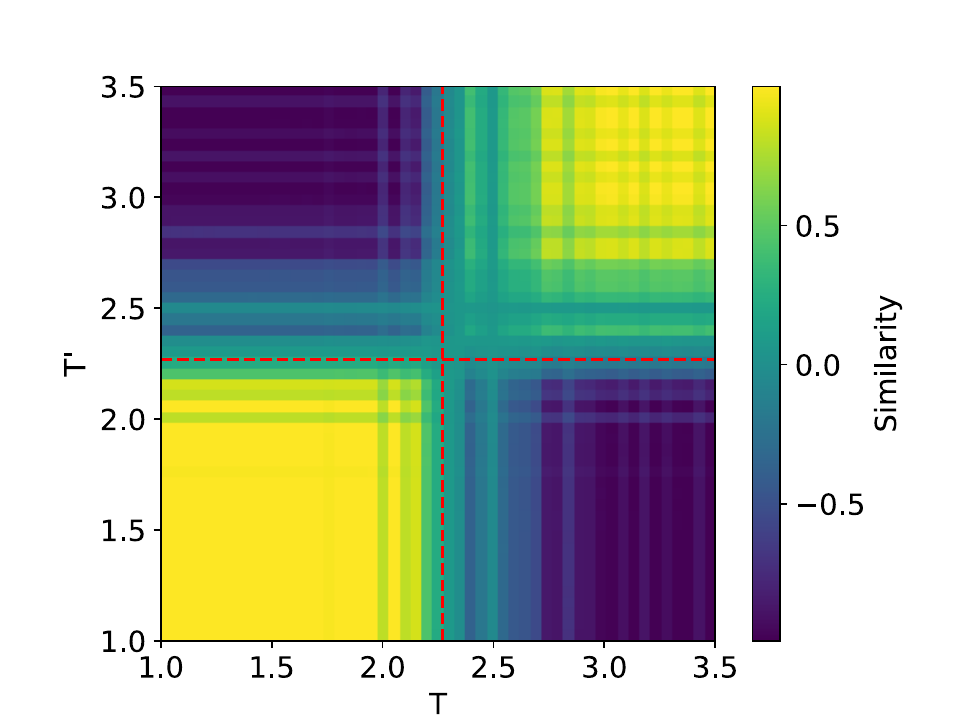}
    \end{minipage}
    \hfill
        \begin{minipage}[t]{0.48\columnwidth}
        \centering
        \includegraphics[width=\linewidth]{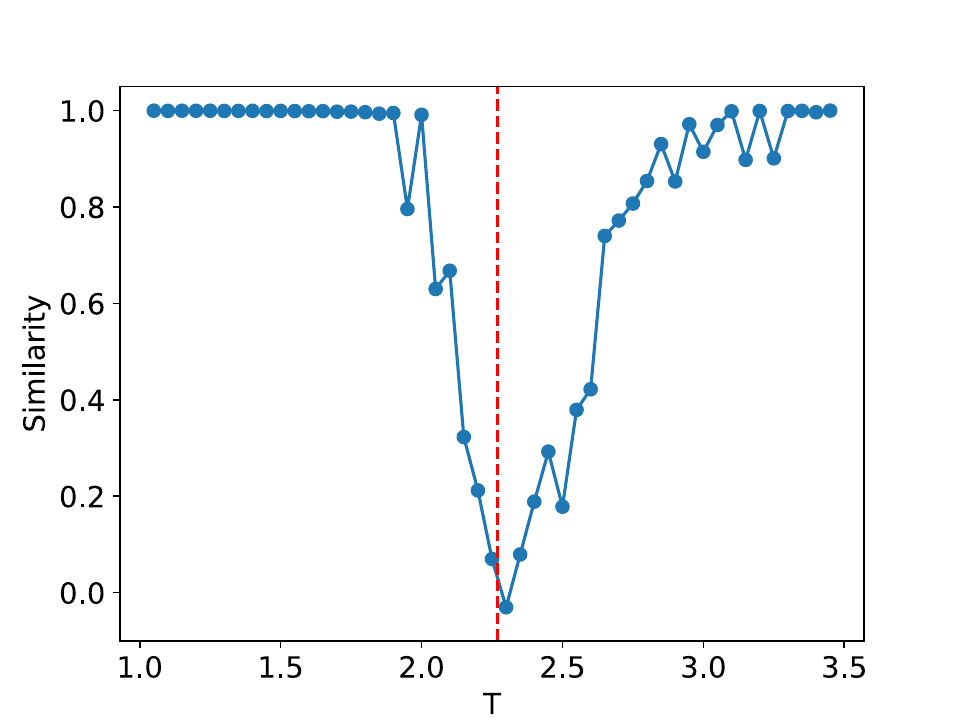}
    \end{minipage}
    \hfill
    \begin{minipage}[t]{0.48\columnwidth}
        \centering
        \includegraphics[width=\linewidth]{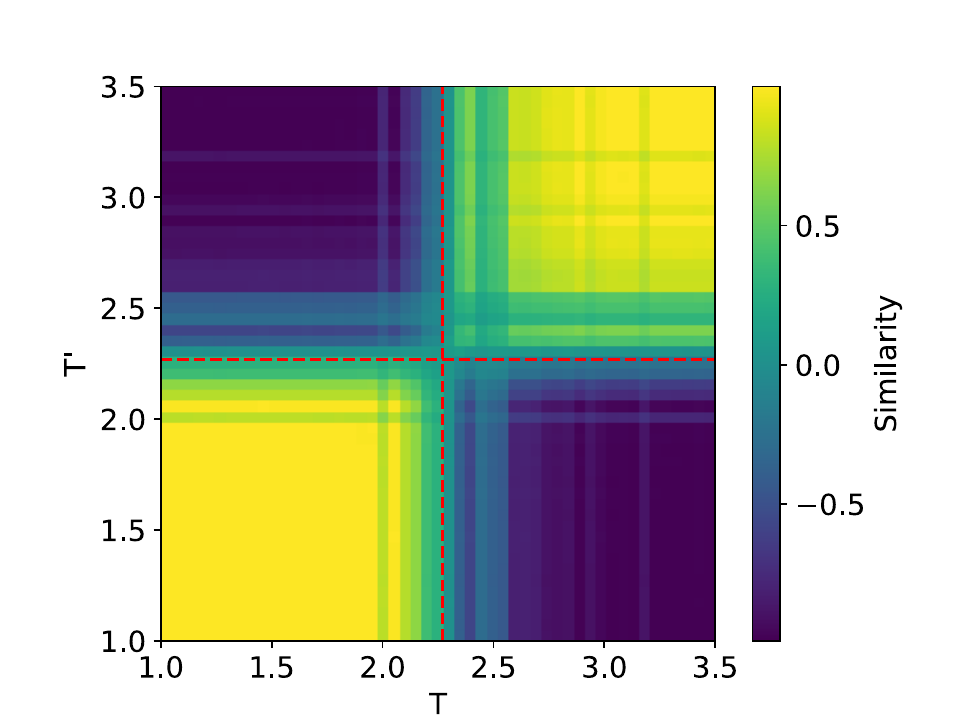}
    \end{minipage}
    \caption{The experimental setup is identical to that of Figure \ref{fig:Ising10-0.0}, except that we use $\sigma=0.3$.}
    \label{fig:Ising10-0.3}
\end{figure}

\section{Additional result for Ising model}
\label{sec:ising}
\begin{figure*}[t]
    \centering
    \begin{minipage}[t]{0.32\textwidth}
        \centering
        \includegraphics[width=\linewidth]{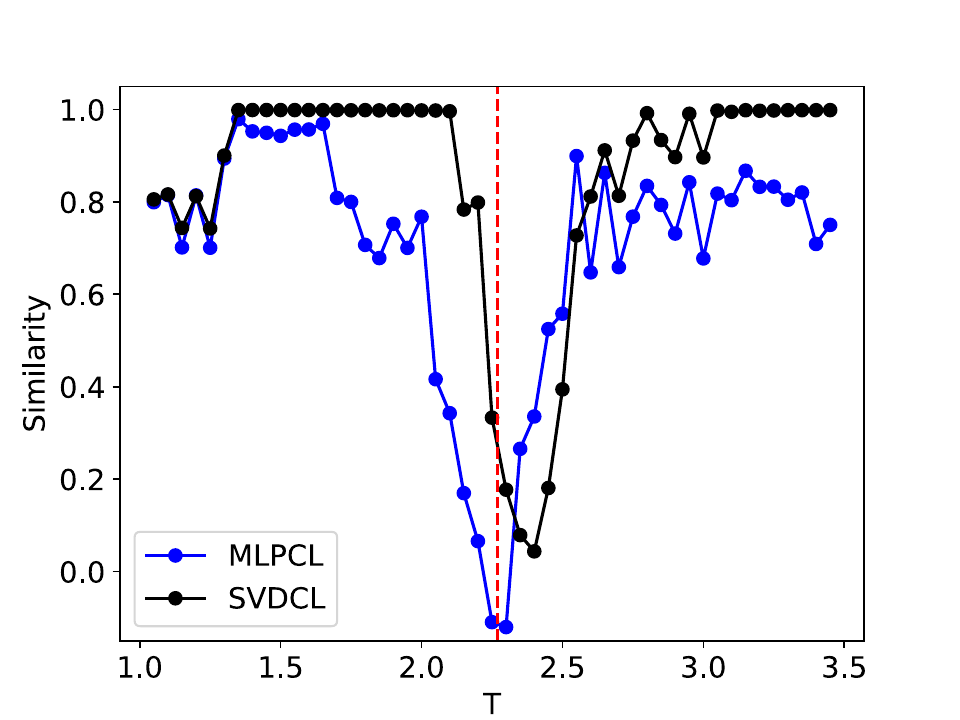}
    \end{minipage}\hfill
    \begin{minipage}[t]{0.32\textwidth}
        \centering
        \includegraphics[width=\linewidth]{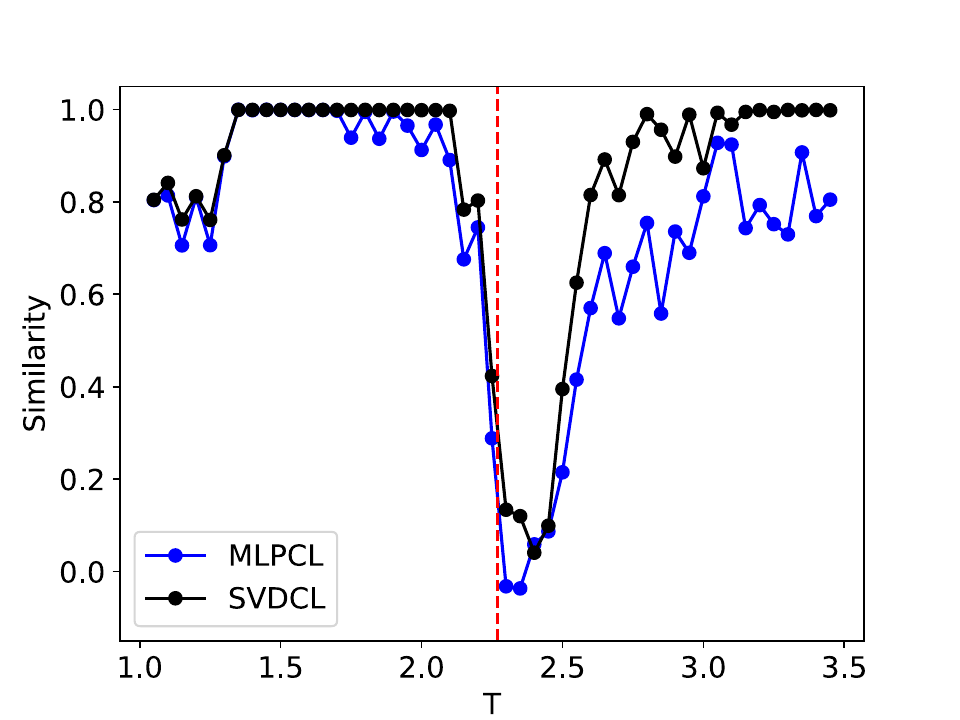}
    \end{minipage}\hfill
    \begin{minipage}[t]{0.32\textwidth}
        \centering
        \includegraphics[width=\linewidth]{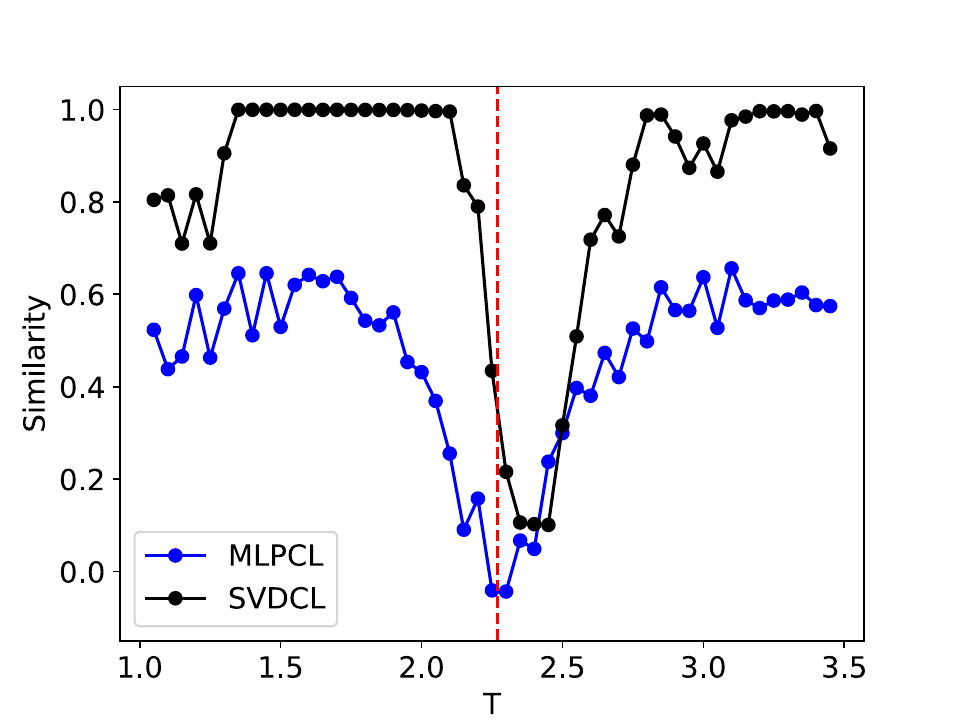}
    \end{minipage}
    \caption{Results for the Ising model with lattice size $L=20$, shown for noise levels 0.0, 0.3, and 0.5 (from left to right).}
    \label{fig:Ising20}
\end{figure*}
An additional experimental result on the Ising model with $L=10$ and $\sigma = 0.3$ is presented in Figure~\ref{fig:Ising10-0.3}.
In this case, aside from our method exhibiting a sharper minimum than MLPCL, the overall behavior is similar to that observed for $\sigma = 0.0$. 
The heatmap still displays a clear separation between the yellow and dark-blue regions. Moreover, our method yields a more distinct boundary than MLPCL, indicating that it is more robust to noise.

For further comparison, we also present results for the Ising model with lattice size $L=20$, shown in Figure~\ref{fig:Ising20}. As the same-scale neural network is employed, the similarity curve appears substantially more irregular in this more complex scenario. Although it uses only 61\% of the parameters, our model maintains consistent performance across noise levels, indicating strong robustness to noise. While its estimate of the critical temperature is somewhat less precise, the out-of-phase similarity produced by MLPCL deteriorates sharply and is heavily corrupted by noise, preventing it from approaching unity. This behavior is consistent with the results observed for the $L=10$ Ising model as $\sigma$ increases from 0.0 to 0.5.

\bibliography{apstemplate}

\end{document}